\documentclass[10pt,twocolumn,letterpaper]{article}

\usepackage[pagenumbers]{iccv} 

\definecolor{iccvblue}{rgb}{0.21,0.49,0.74}
\usepackage[pagebackref,breaklinks,colorlinks,allcolors=iccvblue]{hyperref}

\def\paperID{5706} 
\def\confName{ICCV}
\def\confYear{2025}

\title{Towards More Diverse and Challenging Pre-training for Point Cloud Learning: Self-Supervised Cross Reconstruction with Decoupled Views}

\author{
Xiangdong Zhang$^{*}$, Shaofeng Zhang$^{*}$, Junchi Yan$^{\ddagger}$ \\
School of AI, Shanghai Jiao Tong University
\\
\texttt{\{zhangxiangdong, sherrylone, yanjunchi\}@sjtu.edu.cn}
}

\usepackage{multicol}
\usepackage{multirow}
\usepackage{utfsym}
\usepackage{colortbl}
\usepackage{subcaption}

\newcommand{\mname}{Point-PQAE}
\newcommand{\rightSymbol}{\usym{2713}}
\newcommand{\wrongSymbol}{\usym{2717}}

\begin{document}

\maketitle
\renewcommand{\thefootnote}{\fnsymbol{footnote}}
\footnotetext{$^*$Equal contribution. 
$^\ddagger$Corresponding author. This paper is in part supported by Shanghai Municipal Science and Technology Major Project, China, under grant No.
2021SHZDZX0102.}
\renewcommand{\thefootnote}{\arabic{footnote}}

\begin{abstract}
Point cloud learning, especially in a self-supervised way without manual labels, has gained growing attention in both vision and learning communities due to its potential utility in a wide range of applications. Most existing generative approaches for point cloud self-supervised learning focus on recovering masked points from visible ones within a single view. Recognizing that a two-view pre-training paradigm inherently introduces greater diversity and variance, it may thus enable more challenging and informative pre-training. Inspired by this, we explore the potential of two-view learning in this domain. In this paper, we propose \textbf{Point-PQAE}, a cross-reconstruction generative paradigm that first generates two decoupled point clouds/views and then reconstructs one from the other. To achieve this goal, we develop a crop mechanism for point cloud view generation for the first time and further propose a novel positional encoding to represent the 3D relative position between the two decoupled views. The cross-reconstruction significantly increases the difficulty of pre-training compared to self-reconstruction, which enables our method to surpass previous single-modal self-reconstruction methods in 3D self-supervised learning. Specifically, it outperforms the self-reconstruction baseline (Point-MAE) by 6.5\%, 7.0\%, and 6.7\% in three variants of ScanObjectNN with the \textsc{Mlp-Linear} evaluation protocol. The code is available at \hypersetup{pdfborder={0 0 0}}
\textcolor{magenta}{\url{https://github.com/aHapBean/Point-PQAE}} \hypersetup{pdfborder={0 0 1}}.
\end{abstract}
    
\vspace{-10pt}
\section{Introduction}
\label{sec:intro}

3D vision is gaining increasing attention for its wide applications such as autonomous driving~\cite{qian20223dAutoDriving} and robotics~\cite{sergiyenko20203dRobotics,enebuse2021visionRobot}, 
owing to its ability to understand the human world. The point cloud is the most popular representation form of data in 3D vision and many analyses based on point cloud are explored to solve various tasks such as object classification~\cite{qi2017pointnet,qi2017pointnet++,wang2019DGCNN,li2018pointcnn,guo2021PCT,xu2024towards_PC_representation}, object detection~\cite{misra2021end,qi2019deep,zhou2018voxelnet} and segmentation~\cite{qi2017pointnet,wang2019DGCNN,li2018pointcnn,guo2021PCT,wu2023PTV3}, they typically necessitate fully-supervised training from scratch. Compared to 2D data, point cloud generally requires more expensive and labor-intensive efforts to collect and annotate, which hinders the development of the fully supervised 3D representation learning method. Self-supervised learning is one of the predominant approaches to address this issue and has been proven to be effective in 2D vision~\cite{chen2020SimCLR,he2022VisionMAE,caron2021DINO,zhou2021iBOT,he2020MoCo,grill2020BYOL, zhang2021zero, zhang2020diversified, zhang2022m, zhangcr2pq}. Inspired by this, self-supervised learning has been widely studied in the 3D field in recent years.

\begin{figure}[tb!]
    \centering
    \includegraphics[width=0.9\linewidth]{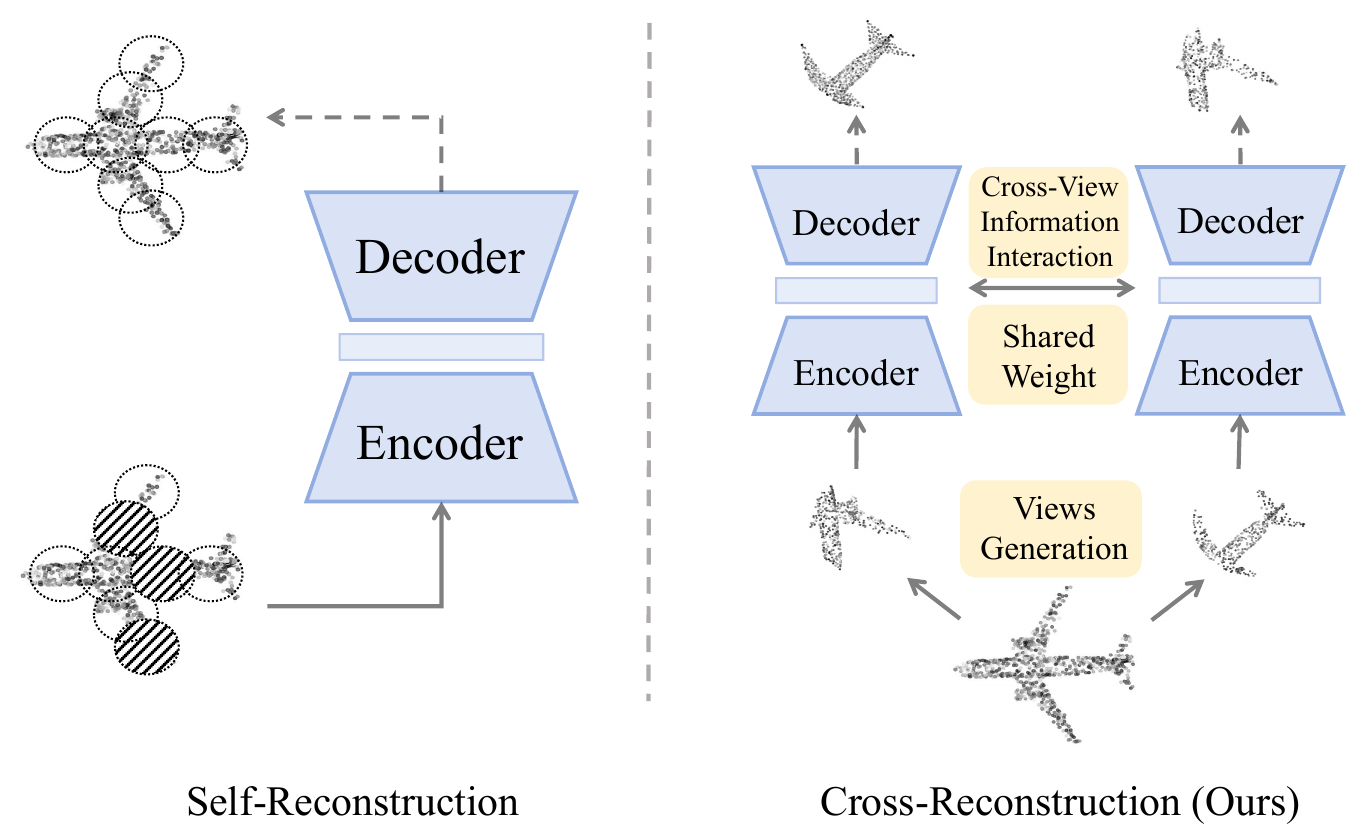}
     \vspace{-7pt}
    \caption{Comparison between self-reconstruction (Point-MAE~\cite{pang2022PointMAE} and other methods based on it) and cross-reconstruction (Ours) paradigms. In the self-reconstruction paradigm, part of the input data is masked, and the autoencoder is trained to recover the missing patches. In contrast, the cross-reconstruction paradigm generates decoupled views first, with one view leveraging cross-view information to reconstruct the other, using an autoencoder backbone.}
    \label{fig:contrast_two_paradigm}
    \vspace{-18pt}
\end{figure}

Self-supervised learning in 3D can be mainly divided into two categories, \ie, generative methods~\cite{chen2024pointGPT,pang2022PointMAE,wang2021OcCo,li2018soNet_GenerativeMethod,achlioptas2018learning_GenerativeMethod,zhang2022pointM2AE,liu2022maskedpoint,zhang2024pcpmae}
and contrastive methods~\cite{afham2022crosspoint,xie2020pointcontrast,dong2022ACT,qi2023Recon,huang2021spatio,sanghi2020info3d,du2021selfcontrastive}. 
Similar to 2D, \textbf{contrastive methods} in 3D aim to learn global-discriminative information by maximizing the mutual information across views~\cite{afham2022crosspoint,xie2020pointcontrast,dong2022ACT} with intra-/inter- modal information~\cite{wu2025mixtureofscores, tan2023datapruneinfomax, 10262344, tan2024saco, tan2024ensembleqap, tan2023movingonesampleout, tan2023semanticdiffusion, tan2021proxygraph, tan2025diffin}. 
PointContrast~\cite{xie2020pointcontrast} is the first to learn transformation invariance across views through the extension of InfoNCE objective~\cite{oord2018infoNCE,xie2020pointcontrast}. 
Different from contrastive methods, the core idea of the \textbf{generative methods}~\cite{pang2022PointMAE,zhang2022pointM2AE,liu2022maskedpoint} in 3D is inspired by MAE~\cite{he2022VisionMAE}, which masks the input point cloud and uses the visible part to reconstruct the masked part. Point-MAE~\cite{pang2022PointMAE} is one of the conventional methods, which learns point-wise information through the self-reconstruction mechanism. Specifically, Point-MAE masks a high ratio of the point cloud and reconstructs the masked points through the visible part. Point-M2AE~\cite{zhang2022pointM2AE} proposes a hierarchical Transformer~\cite{liu2021swinTransformer} with a corresponding masking strategy based on it, capturing both fine-grained and high-level semantics of 3D shapes.

\begin{figure*}[tb!]
    \centering
    \includegraphics[width=0.95\textwidth]{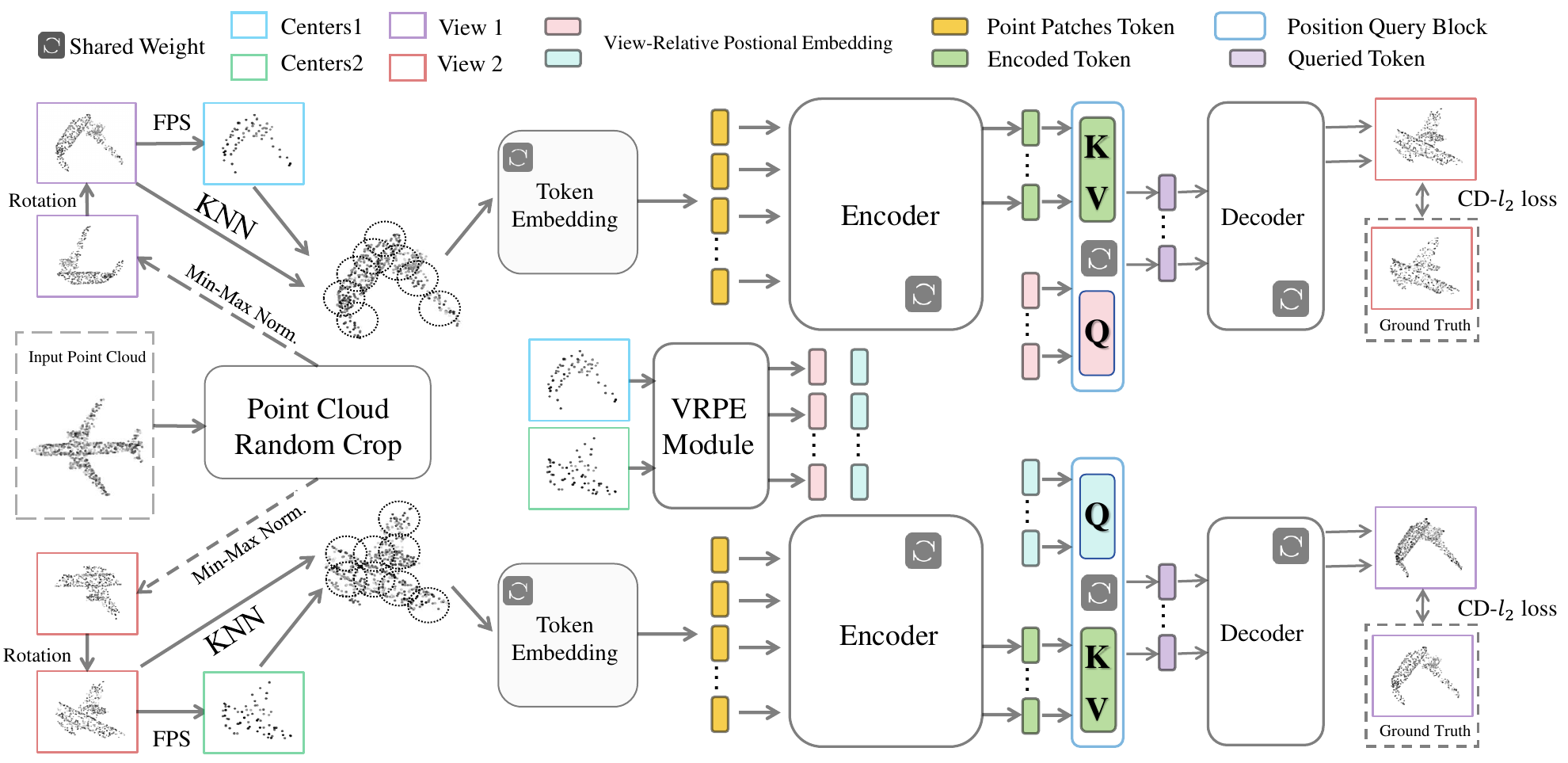}
    \caption{Pipeline of Point-PQAE. The input point cloud is randomly cropped followed by the rotation to generate views. Then, we feed the views to the patch embedding layer and the transformer encoder, followed by the proposed positional query block. The View-Relative Positional Embedding (VRPE) is obtained through the VRPE module by extracting relative geometric relations, which is taken as ``Query'' in the cross-attention mechanism. The queried hidden embeddings are then fed to the decoder to predict the inputs of the other view.}
    \label{fig:overview}
    \vspace{-15pt}
\end{figure*}

Existing generative methods mostly focus on the masked reconstruction in a single view, where single-view learning has been proven less difficult and informative than two views~\cite{chen2020SimCLR, grill2020BYOL} since two views would bring more variance than a single view. To bring the benefits of the two-view learning paradigm to the generative learning field, one straightforward assumption is adopting the two-view learning paradigm in the 3D generative pertaining tasks to increase the difficulty of point-wise reconstruction. However, unlike 2D vision data, constructing two views in the 3D point cloud is more challenging, as point cloud data is less structured. Besides, cross-reconstruction (reconstructing one view from the other view) is much more difficult than self-reconstruction (reconstructing the masked single view), 
which involves learning both the inter-positional relation between two views and the inner spatial information within the view. The differences between these two paradigms can be seen in~\cref{fig:contrast_two_paradigm}. Aimed at solving the above issues, we propose {\mname} (\textbf{P}osition \textbf{Q}uery based \textbf{A}uto\textbf{E}ncoder for \textbf{Point} Cloud), a novel framework aimed at addressing two primary challenges:
i) how to construct two views in point cloud data, and ii) how to perform cross-reconstruction between two views. 
As shown in~\cref{fig:overview}, for each point cloud, we first apply a custom-designed point cloud crop mechanism, which randomly selects two points as the central points of two views and records their positions. Following this, for each central point, the nearest points according to a specified ratio are incorporated into the view. Subsequently, we normalize the cropped point cloud and apply random augmentation (\eg, rotation) to each point cloud to achieve view decoupling. Then we calculate the view-relative positional embedding (VRPE) for these two decoupled views. Finally, we take the VRPE and one view of the point cloud as input to predict the point-wise input of the other view. {\textbf{The highlights of this paper are:}}

\noindent i) \textbf{New generative framework:} We propose {\mname}, the first framework that successfully brings cross-reconstruction to 3D generative field for point cloud self-supervised learning, with three modules: 1) decoupled views generation, 2) VRPE generation, and 3) positional query block. To our knowledge, we are the first to design and apply crop mechanism to point cloud self-supervised learning. Our framework, Point-PQAE, enables more informative and challenging self-supervised pre-training compared to existing self-reconstruction methods. 

\noindent ii) {\textbf{The first relative position-aware query module for point cloud:}} We introduce a positional query block after the encoder to capture relative position information between decoupled views. The module applies cross-attention between the hidden representations of one view and the VRPE, enabling it to predict the other view. Our module can be easily plugged into various related tasks, \eg, distillation as stated in Appendix Sec. 4.

\noindent iii) {\textbf{Strong performance:}} 
    {\mname} achieves new state-of-the-art performance on several benchmarks, \eg, outperforming previous methods on few-shot learning and achieving average improvements of 1.8\%, 6.7\%, and 4.4\% over the baseline (Point-MAE) on ScanObjectNN classification across three evaluation protocols (\textsc{Full}, \textsc{Mlp-Linear}, \textsc{Mlp-3}), respectively.

\section{Related works}
\label{sec:related_works}
\noindent \textbf{SSL in 2D vision.} 
Self-supervised learning (SSL) in 2D vision is broadly categorized into contrastive-based and masked-image-modeling (MIM) approaches. Contrastive methods learn discriminative features by creating two augmented views of an image and enforcing invariance between them. For instance, SimCLR~\cite{chen2020SimCLR} treats augmented views from the same image as positive pairs and those from different images as negatives. MoCo~\cite{he2020MoCo} utilizes a memory bank to store these negative samples and further proposes a momentum-updated mechanism to avoid model collapse. MIM-based methods mask portions of the input and task models with reconstructing the masked regions, encouraging fine-grained understanding. MAE~\cite{he2022VisionMAE} employs an encoder-decoder architecture, discarding masked patches before reconstruction, while SimMIM~\cite{xie2022simmim} replaces masked regions with a learnable parameter. Subsequent work explores variations such as masking strategies~\cite{zhang2022contextual, li2022semmae} and feature-level reconstruction~\cite{zhou2021iBOT}.



\noindent \textbf{Representation learning for point clouds.} It is a challenging task due to its irregular and sparse nature when compared to other modalities such as images which are well-structured and uniform. 
PointContrast~\cite{xie2020pointcontrast} firstly borrows the idea of contrastive learning in 2D and performs point-level invariant mapping on two transformed views of the given point cloud. On the basis of PointContrast, CrossPoint~\cite{afham2022crosspoint} further proposes inter and intra-modal contrastive objectives to enhance the global-discriminative capability. To better model multi-modal data, ACT~\cite{dong2022ACT} employs cross-modal auto-encoders as teacher models to acquire knowledge from other modalities. In contrast to contrastive objectives, Point-MAE~\cite{pang2022PointMAE} proposes to randomly mask points, and reconstruct the masked part through the visible part. Following Point-MAE and CrossPoint, ReCon~\cite{qi2023Recon} harmoniously incorporates the generative method Point-MAE and the contrastive method. \textbf{Almost all the existing generative methods in 3D rely on self-reconstruction.} 
To this end, we propose a new cross-reconstruction generative framework, named {\mname}, which increases pre-training difficulty by introducing cross-view reconstruction, bringing more variance to the training samples compared to single-view reconstruction. This requires the model to effectively learn both intra-view and inter-view knowledge, thereby enabling the model to learn more semantic representations.
\section{The proposed Point-PQAE}

\subsection{Views and embeddings generation}
\vspace{-4pt}
{\bf Decoupled views generation.} 
\label{sec:view-generation}
Random Image Crop has been widely used in both the supervised domain~\cite{he2016resNet,simonyan2014vgg,krizhevsky2012AlexNet} and the self-supervised domain~\cite{chen2020SimCLR, he2020MoCo, bao2021beit}, and it has been proved to be effective in practice. This method typically involves randomly selecting a rectangular region from an image and cropping it. However, to the best of our knowledge, no corresponding crop mechanism has been proposed for 3D self-supervised representation learning. Instead of directly applying the 2D crop algorithm to 3D (by randomly selecting a cube in space), which may lead to inconsistency, as the points within cubes of the same size can vary significantly due to the diverse distribution of points in 3D, we design in a smarter way: For a randomly selected crop ratio, we first select a center point randomly and then choose the corresponding number of points in the point cloud that are closest to the center point. 

Formally, given an input point cloud with p points $\mathbf{X} \in \mathbb{R}^{p\times 3}$ and a pre-defined minimum crop ratio $r_m$ where $0<r_m<1$, we first randomly select the crop ratios $r_1$ and $r_2$ uniformly from $\left[r_m,1\right]$. Then, we randomly select two center points $\mathbf{C_1}, \mathbf{C_2} \in \mathbb{R}^{1\times3}$ and get $\mathbf{X}_1$ and $\mathbf{X}_2$ by expanding nearest $r_1\times p$ and $r_2\times p$ points to the center point $\mathbf{C_1}$ and $\mathbf{C_2}$ respectively in $\mathbf{X}$. Meanwhile, we record the absolute coordinates of cropped point clouds' geometric centers, which are denoted as $\mathbf{L}_1 \in \mathbb{R}^{1\times3}$ and $\mathbf{L}_2 \in \mathbb{R}^{1\times3}$, respectively. The process can be formulated as:
\begin{align}
    r_1, r_2 &= \operatorname{UniformSample}([r_m, 1]) \\
    \mathbf{C_1}, \mathbf{C_2} &= \operatorname{RandomSelect}(\mathbf{X}) \\
    \mathbf{X}_1 &= \operatorname{N\text{-}Nearest}(\mathbf{C_1}, r_1, \mathbf{X}) \\
    \mathbf{X}_2 &= \operatorname{N\text{-}Nearest}(\mathbf{C_2}, r_2, \mathbf{X})
\end{align}
where $\mathbf{X}_1 \in \mathbb{R}^{r_1 p \times 3}$ and $\mathbf{X}_2 \in \mathbb{R}^{r_2 p \times 3}$ mean the two cropped views. The $\mathbf{X}_1$ and $\mathbf{X}_2$ are normalized using the min-max normalization, {centered on $\mathbf{L_1}$ and $\mathbf{L_2}$,} respectively. Subsequently, we apply additional augmentation to further enhance the variance between two views. We apply random rotations to the cropped views. The distinct normalization centers, along with random rotation, effectively isolate the coordinate systems of the two views and alter their fixed relative relationships. This process decouples the two point clouds, resulting in two independent views.


{\bf Point patches generation.} 
We divide the input point cloud into irregular point patches (may overlap) via the Farthest Point Sampling(FPS) and K-Nearest Neighborhood (KNN) algorithm following Point-BERT~\cite{yu2022pointBert}. Formally, given two point clouds (views) $\mathbf{X}_1$ and $\mathbf{X}_2$, we first use the FPS algorithm to sample pre-defined $n$ points as group centers, which we called $\mathbf{G}_1 \in \mathbb{R}^{n\times 3}$ and $\mathbf{G}_2 \in \mathbb{R}^{n\times 3}$, respectively. Then, based on the center points, we use KNN to choose the pre-defined $k$-nearest points for each group center $\mathbf{G}^{(i)}\ (0 \leq i \leq n)$ and finally get $n$ groups (each group contains $k$ points). We apply the FPS and KNN on $\mathbf{X}_1$ and $\mathbf{X}_2$, respectively:
\begin{align} 
    \label{eqn:FPS}   
    \mathbf{G}_1 =\operatorname{FPS}(\mathbf{X}_1), & \ \ \ \mathbf{G}_2=\operatorname{FPS}(\mathbf{X}_2) \\
    \label{eqn:KNN}
    \mathbf{P}_1=\operatorname{KNN}(\mathbf{X}_1,\mathbf{G}_1), & \ \ \ \mathbf{P}_2=\operatorname{KNN}(\mathbf{X}_2,\mathbf{G}_2)
\end{align}
where $\mathbf{G}_1,\ \mathbf{G}_2 \in \mathbb{R}^{n\times 3}$ and $\mathbf{P}_1, \ \mathbf{P}_2 \in \mathbb{R}^{n\times (k \cdot 3)}$. Then, the $\mathbf{P}_1$ and $\mathbf{P}_2$ both are treated as a sequence of length $n$ and dimension $k \cdot 3$ as the input.


\vspace{-4pt}
\subsection{Backbone with positional query}
\vspace{-2pt}
\label{pq_method}
Following~\cite{pang2022PointMAE}, we adopt a lightweight PointNet~\cite{qi2017pointnet}, followed by an asymmetric encoder-decoder structure consisting of Transformer blocks in the {\mname}. After the encoder projects the input embedding into the latent space, we use the latent representation of one view and view-relative positional embedding (VRPE) between two views to implement the position query through cross-attention~\cite{vaswani2017attentionIsallyouneed}. We directly feed the output of the position query block to the decoder to reconstruct another view without using extra positional embedding. The last layer of the model adopts a simple prediction head to achieve the cross-reconstruction.

{\bf Encoding.} 
The encoder consists of the lightweight PointNet and standard Transformer blocks~\cite{vaswani2017attentionIsallyouneed}, which projects $\mathbf{P}_1$ and $\mathbf{P}_2$ to latent space, respectively:
\begin{align}
    \mathbf{H}_1=\operatorname{Encoder}(\mathbf{P}_1, \mathbf{G}_1), \ \ \mathbf{H}_2=\operatorname{Encoder}(\mathbf{P}_2, \mathbf{G}_2),
\end{align}
where $\mathbf{H}_1 \in \mathbb{R}^{n\times D}$ and $\mathbf{H}_2 \in \mathbb{R}^{n\times D}$ mean the latent representations of the view 1 and view 2. Note that $D$ is the hidden dimension of the networks.

{\bf Positional query block.} 
\label{sec:pq}
After obtaining latent representations $\mathbf{H}_1$ and $\mathbf{H}_2$, we feed them into the proposed positional query block to extract cross-information. The module can be mainly divided into 3 sub-stages: attain relative information, VRPE generation, and cross-attention.

{\it Attain relative information.} 
Take reconstructing view 2 from view 1 as an example. To reconstruct view 2, we have to know the latent representations of view 1 and the relative positions of view 2 centered on view 1, where the latent representations of view 1 can be obtained through the encoder network. Therefore, the key to reconstructing view 2 is to obtain the relative positions of view 2 centered on view 1. Back to the crop operation, we have recorded the geometric centers of views, denoted as $\mathbf{L}_1$ and $\mathbf{L}_2$, respectively. We define the relative position of view 2 centered on view 1 as $\mathbf{RL}_{1\rightarrow 2} = \mathbf{L}_1 - \mathbf{L}_2$ where $\mathbf{RL}_{1\rightarrow 2} \in \mathbb{R}^{1\times 3}$.
To better align the dimension in the latter operation, we expand the $\mathbf{RL}_{1\rightarrow 2}$ to $\mathbb{R}^{n\times 3}$ by repeating the content for $n$ times. By obtaining the relative positions between the two views, we further add the group center location to help the PQ modules reconstruct each group of the inputs. Specifically, we define the group-/patch- wise relative positions as:
\begin{align}
    \mathbf{RP}_{1\rightarrow 2} &=\operatorname{ConCat}(\mathbf{G}_2, \mathbf{RL}_{1\rightarrow 2}),\ \ \\
    \mathbf{RP}_{2\rightarrow 1} &=\operatorname{ConCat}(\mathbf{G}_1, \mathbf{RL}_{2\rightarrow 1})
\end{align}
where $\mathbf{RP} \in \mathbb{R}^{n\times 6}$ (as $\mathbf{G} \in \mathbb{R}^{n\times 3}$ and $\mathbf{RL} \in \mathbb{R}^{n\times 3}$) and $\operatorname{ConCat}(\cdot, \cdot)$ is the concatenate operation.

{\it View-relative positional embedding (VRPE).}  
Instead of simply using learnable positional encoding, which could hurt the expression of the relative positions between the two views (see ablation in Sec.~\ref{sec:abla_embedding}), we use the fixed positional encoding of relative information to cut the uncertainty. Specifically, for elements in the second dimension of $\mathbf{RP}_{1\rightarrow 2}$, we use a sinusoid form~\cite{he2022VisionMAE} in Transformers to generate view-relative positional embedding $\mathbf{VRPE}_{1\rightarrow 2}$:
\begin{align}
\label{eq:sin-cos-1}
\mathbf{VRPE}_{1\rightarrow 2}^{i} = &\left[\sin\left(\frac{\mathbf{RP}^{i}_{1\rightarrow 2}}{e^{2\times \frac{1}{D/12}}}\right),\cos\left(\frac{\mathbf{RP}^{i}_{1\rightarrow 2}}{e^{2\times \frac{1}{D/12}}}\right), \right. \nonumber \\
&\left.\sin\left(\frac{\mathbf{RP}^{i}_{1\rightarrow 2}}{e^{2\times \frac{2}{D/12}}}\right), \cos\left(\frac{\mathbf{RP}^{i}_{1\rightarrow 2}}{e^{2\times \frac{2}{D/12}}}\right), \right. \dots, \nonumber \\
&\left.\sin\left(\frac{\mathbf{RP}^{i}_{1\rightarrow 2}}{e}\right), \cos\left(\frac{\mathbf{RP}^{i}_{1\rightarrow 2}}{e}\right) \right]
\end{align}

where $\mathbf{RP}^i_{1\rightarrow2} \in \mathbb{R}^{n\times1}$, $\mathbf{VRPE}^{i}_{1\rightarrow 2} \in \mathbb{R}^{n\times (D/6)}\ (1 \leq i \leq 6)$. $e = 10000$ is pre-defined, which is also used in MAE~\cite{he2022VisionMAE}. Finally, we concatenate $\mathbf{VRPE}^{i}_{1\rightarrow 2}$ for each dimension to obtain $\mathbf{VRPE}_{1\rightarrow 2} \in \mathbb{R}^{n\times D}$. 

{\it Cross-attention.} 
Empirically, assume the latent representations of view 1, denoted as $\mathbf{H}_1$, contains intrinsic features and the global characteristics of its source point cloud, then with the combination of the view-relative positional embeddings $\mathbf{VRPE}_{1\rightarrow 2}$, the view 2 can be reconstructed. Specifically, we employ $\mathbf{VRPE}_{1\rightarrow 2}$ (as $\mathbf{Q}$) and $\mathbf{H}_1$ (as $\mathbf{K}$, $\mathbf{V}$) in cross-attention mechanisms, as can be formalized:
\begin{align}
    \mathbf{T}_2 &= \operatorname{Attn}(\mathbf{Q}_{\mathbf{VRPE}_{1\rightarrow 2}},\mathbf{K}_{\mathbf{H}_1}, \mathbf{V}_{\mathbf{H}_1}) \notag \\
    &= \operatorname{Softmax}\left(\frac{\mathbf{Q}_{\mathbf{VRPE}_{1\rightarrow 2}}\mathbf{K}_{\mathbf{H}_1}}{\sqrt{D}}\right)\mathbf{V}_{\mathbf{H}_1}
\end{align}
where $\mathbf{Q}_{\mathbf{VRPE}_{1\rightarrow 2}} = \mathbf{VRPE}_{1\rightarrow 2}\mathbf{W}_{\mathbf{Q}}$, $\mathbf{K}_{\mathbf{H}_1} = \mathbf{H}_{1} \mathbf{W}_{\mathbf{K}}$, and $\mathbf{V}_{\mathbf{H}_1} = \mathbf{H}_{1} \mathbf{W}_{\mathbf{V}}$, where $\mathbf{W}_{\mathbf{Q}}$, $\mathbf{W}_{\mathbf{K}}$, and $\mathbf{W}_{\mathbf{V}}$ are learnable parameters. Reconstructing view 1 from view 2 is a siamese condition of this, through the same way as stated above, we can obtain $\mathbf{RL}_{2\rightarrow 1}$, $\mathbf{RP}_{2\rightarrow 1}$, $\mathbf{VRPE}_{2\rightarrow 1}$ and yield $\mathbf{T}_1$. 

{\bf Decoding.} The obtained $\mathbf{T}_1$ and $\mathbf{T}_2$ are fed into a decoder composed of a few transformer blocks:
\begin{equation}
    \mathbf{Z}_1=\operatorname{Decoder}(\mathbf{T}_1), \ \ \ \mathbf{Z}_2=\operatorname{Decoder}(\mathbf{T}_2)
\end{equation}
Finally, similar to~\cite{pang2022PointMAE}, we adopt a projection head (composed of a fully connected layer) to predict the input point cloud as follows,  where $\mathbf{P}^1_{pred},\mathbf{P}^2_{pred} \in \mathbb{R}^{n\times k \times 3}$: 
\begin{align}
    \mathbf{P}^1_{pred} &=\operatorname{Reshape}(\operatorname{Linear}(\mathbf{Z}_1)), \\ \mathbf{P}^2_{pred} &=\operatorname{Reshape}(\operatorname{Linear}(\mathbf{Z}_2))
\end{align}

\vspace{-6pt}
\subsection{Objective function}
\vspace{-4pt}
\label{sec:siamese-loss}
The overall objective is to perform cross-reconstruction, which involves predicting view 1 from view 2 and predicting view 2 from view 1. Given the predicted point patches $\mathbf{P}_{pred}^1$ and $\mathbf{P}_{pred}^2$ and ground truth $\mathbf{P}_1$ and $\mathbf{P}_2$, we compute the cross-reconstruction loss using the $l_2$ Chamfer Distance loss function~\cite{fan2017chamferDist}, written as $\mathcal{L}_{cross}=\mathcal{L}_{2\rightarrow 1}+\mathcal{L}_{1\rightarrow 2}$, where: 
\begin{equation}
\begin{aligned}
    \mathcal{L}_{2\rightarrow 1}= &\frac{1}{\left|\mathbf{P}_{pred}^1\right|}\sum_{a\in \mathbf{P}_{pred}^1} \min_{b\in \mathbf{P}_1} \|a - b\|^2_2 \\
    &+ \frac{1}{\left|\mathbf{P}_1\right|}\sum_{b\in \mathbf{P}_1} \min_{a\in \mathbf{P}_{pred}^1} \|a - b\|^2_2 
\end{aligned}    
\end{equation}
where $\left|\mathbf{P} \right|$ is the cardinality of the set $\mathbf{P}$ and $\|\cdot\|_2$ is the $l_2$ distance. The $\mathcal{L}_{1\rightarrow 2}$ follows similarly.
\section{Experiments}

This section is organized as follows: First, we present the model architecture and pre-training details. Second, we validate the effectiveness of our pre-trained model on a wide range of downstream tasks, including object classification, few-shot learning, and part segmentation. Finally, ablation studies are conducted to demonstrate the properties and robustness of our proposed Point-PQAE.
\vspace{-4pt}
\subsection{Pre-training setups}
\vspace{-2pt}
{\bf Pre-trained dataset.} 
We use the dataset ShapeNet~\cite{chang2015ShapeNet} for pre-training of Point-PQAE following previous studies~\cite{pang2022PointMAE,dong2022ACT,qi2023Recon}. ShapeNet~\cite{chang2015ShapeNet} consists of about 51,300 clean 3D models, covering 55 common object categories.

{\bf Model structure.} 
We adopt standard Transformer blocks~\cite{vaswani2017attentionIsallyouneed} in the autoencoder's backbone where the encoder has 12 Transformer blocks and the decoder has 4 blocks. MLP ratio in Transformer blocks is set to 4. Each Transformer block has 384 hidden dimensions and 6 heads. The positional query module between encoder and decoder only performs cross-attention mechanism once as stated in~\cref{pq_method}, and the number of heads is set to 6 and remains 384 hidden dimensions.

\begin{figure*}[tb!]
    \centering
    \includegraphics[width=0.8\textwidth]{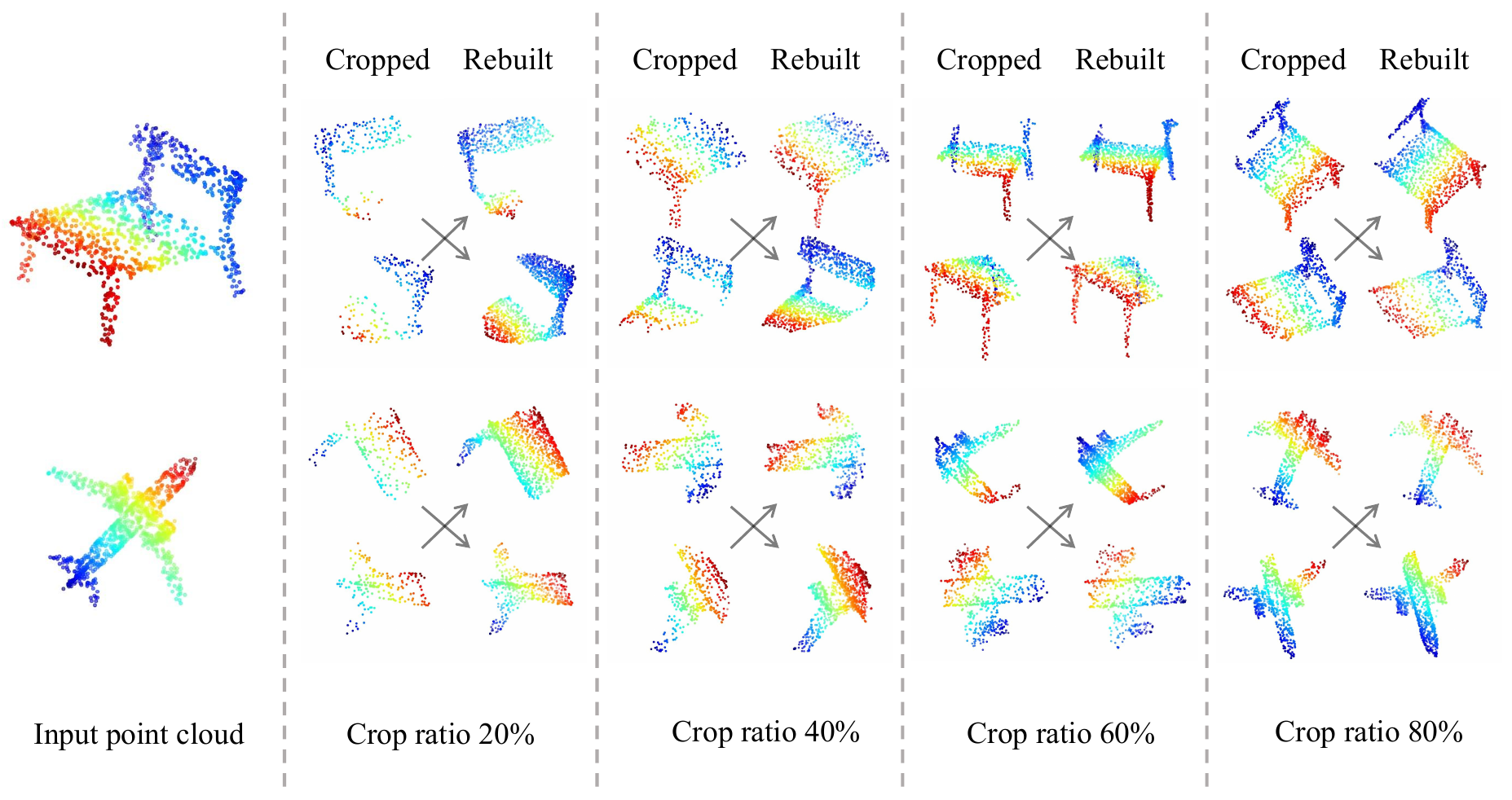}
    \vspace{-10pt}
    \caption{Cross-reconstruction results on ShapeNet. The arrow points from source point clouds to cross-reconstruction results. Point-PQAE generalizes well to other crop ratios though with minimum crop ratio $r_m=0.6$ when pre-training.}
    \label{fig:pretrain_result}
    \vspace{-15pt}
\end{figure*}

{\bf Experiment detail.} 
For each input point cloud, we only apply random rotation for pre-training data augmentation. After sampling 1024 points via FPS from the input point cloud, we generate two different partial point clouds by randomly cropping and rotating as illustrated in~\cref{sec:view-generation} with $r_m=0.6$. 
Both of them are divided into 64 patches with 32 points via FPS and KNN. The Point-PQAE model undergoes pre-training for 300 epochs using an AdamW~\cite{loshchilov2017AdamWOptimizer} optimizer with a batch size of 128. The initial learning rate is set to 0.0005, with a weight decay of 0.05, with cosine learning rate decay~\cite{loshchilov2016cosLrDecay} utilized. We visualize the pre-training results in~\cref{fig:pretrain_result}. It shows Point-PQAE learns cross-knowledge well and is able to generalize excellently to other crop ratios, even though it was pre-trained with $r_m=0.6$.

\vspace{-4pt}
\subsection{Transfer learning on downstream tasks}
\vspace{-2pt}

{\bf Transfer protocol.} 
We use three transfer learning protocols for classification tasks following~\cite{dong2022ACT}: \textsc{Full}, \textsc{Mlp-Linear}, \textsc{Mlp-3}, which are detailed in Appendix Sec. 3.

{\bf 3D real-world object classification.} 
\label{real_world_cls}
The transferability of models can be well demonstrated through testing on a 3D real-world object dataset. Therefore, we transfer our pre-trained model to ScanObjectNN~\cite{uy2019ScanObjectNN} for classification, which is one of the most challenging 3D datasets, covering approximately 15,000 real-world objects across 15 categories. We conduct experiments on three variants: OBJ-BG, OBJ-ONLY, and PB-T50-RS. The results are reported in~\cref{tab:scanobjnet-modelnet-cls}. Our Point-PQAE surpasses previous single-modal self-supervised methods by a margin under all protocols. In contrast to the most relevant baseline, Point-MAE, Point-PQAE significantly outperforms it, showing improvements of \textbf{2.4\%}, \textbf{1.7\%}, and \textbf{1.2\%}. Especially for the \textsc{Mlp-Linear} and \textsc{Mlp-3} protocols, our Point-PQAE outperforms Point-MAE by \textbf{6.7\%} and \textbf{4.4\%} on average, respectively. The excellent frozen representation evaluation results indicate the superior quality of the representation learned by our Point-PQAE during pre-training. This shows that the cross-reconstruction paradigm learns more robust representations and generalizes better on real-world datasets. 

\begin{table*}[tb!]
    \centering
    \caption{ Classification accuracy (\%) on ScanObjectNN and ModelNet40. The inference model parameters \#P (M) are reported. Three variants are evaluated on ScanObjectNN and the accuracy obtained on ModelNet40 is reported for both 1K and 8K points. We compare methods using the \textcolor{blue}{$\bullet$} plain Transformer architectures, \eg Point-MAE~\cite{pang2022PointMAE}, \textcolor{blue}{$\circ$} hierarchical Transformer architectures and \textcolor{orange}{$\circ$} dedicated architectures for 3D. The dagger($^\dagger$) denotes the baseline results reported from ReCon~\cite{qi2023Recon} which aligns augmentation with us and ReCon.}
    \vspace{-5pt}
    \resizebox{0.7\textwidth}{!}{
    \begin{tabular}{lcccccc}
    \toprule[1pt]
    \multirow{2}{*}{Methods} & \multirow{2}{*}{\#P} & \multicolumn{3}{c}{ScanObjectNN} & \multicolumn{2}{c}{ModelNet40} \\
     & & OBJ\underline{~~}BG & OBJ\underline{~~}ONLY & PB\underline{~~}T50\underline{~~}RS & 1K P & 8K P \\
    \midrule 
    \multicolumn{7}{c}{$Supervised \ Learning \ Only$} \\
    \midrule
    \textcolor{orange}{$\circ$}PointNet~\cite{qi2017pointnet} & 3.5 & 73.3 & 79.2 & 68.0 & 89.2 & 90.8 \\
    \textcolor{orange}{$\circ$}PointNet$++$~\cite{qi2017pointnet++} & 1.5 & 82.3 & 84.3 & 77.9 & 90.7 & 91.9 \\
    \textcolor{orange}{$\circ$}DGCNN~\cite{wang2019DGCNN} & 1.8 & 82.8 & 86.2 & 78.1 & 92.9 & - \\
    \textcolor{orange}{$\circ$}PointCNN~\cite{li2018pointcnn} & 0.6 & 86.1 & 85.5 & 78.5 & 92.2 & - \\
    \textcolor{orange}{$\circ$}SimpleView~\cite{goyal2021simpleView} & - & - & - & 80.5$\pm$0.3 & 93.9 & - \\
    \textcolor{blue}{$\circ$}PCT~\cite{guo2021PCT} & 2.88 & - & - & - & 93.2 & - \\
    \midrule
    \multicolumn{7}{c}{$with \ Self\text{-}Supervised \ Representation \ Learning$ (\textsc{Full})} \\
    \midrule 
    \textcolor{blue}{$\bullet$}Transformer~\cite{vaswani2017attentionIsallyouneed} & 22.1 & 83.0 & 84.0 & 79.1 & 91.4 & 91.8 \\
    \textcolor{blue}{$\bullet$}Point-BERT~\cite{yu2022pointBert} & 22.1 &  87.4 & 88.1 & 83.0 & 93.2 & 93.8 \\
    \textcolor{blue}{$\bullet$}MaskPoint~\cite{liu2022maskedpoint} & - & 89.3 & 88.1 & 84.3 & 93.8 & - \\
    \textcolor{blue}{$\bullet$}Point-MAE~\cite{pang2022PointMAE} & 22.1 & 90.0 & 88.3 & 85.2 & 93.8 & 94.0 \\ 
    \textcolor{blue}{$\circ$}Point-M2AE~\cite{zhang2022pointM2AE} & 15.3 & 91.2 & 88.9 & 86.4 & \textbf{94.0} & - \\
    \textcolor{blue}{$\bullet$}PointDif~\cite{zheng2024pointDif} & 22.1 & 93.3 & 91.9 & 87.6 & - & - \\
    \textcolor{blue}{$\bullet$}Point-MAE$^\dagger$~\cite{pang2022PointMAE} & 22.1 & 92.6 & 91.9 & 88.4 & 93.8 & 94.0 \\ 
    \textcolor{blue}{$\bullet$}\textbf{Point-PQAE} & {22.1} & \textbf{95.0} & \textbf{93.6} & \textbf{89.6} & \textbf{94.0} & {\textbf{94.3}} \\
    \midrule
    \multicolumn{7}{l}{Methods using cross-modal information and teacher models} \\
    \textcolor{blue}{$\bullet$}Joint-MAE~\cite{guo2023jointMAE} & - & 90.9 & 88.9 & 86.1 & 94.0 & - \\
    \textcolor{blue}{$\bullet$}TAP~\cite{wang2023TAP} & 22.1 & 90.4 & 89.5 & 85.7 & 94.0 & - \\
    \textcolor{blue}{$\bullet$}ACT~\cite{dong2022ACT} & 22.1 & 93.3 & 91.9 & 88.2 & 93.7 & 94.0 \\
    \textcolor{orange}{$\circ$}PointMLP+ULIP~\cite{xue2023ulip} & - & - & - & 89.4 & 94.5 & 94.7 \\
    \textcolor{blue}{$\circ$}I2P-MAE~\cite{zhang2023I2PMAE} & 15.3 & 94.2 & 91.6 & 90.1 & 94.1 & \\
    \textcolor{blue}{$\bullet$}ReCon~\cite{qi2023Recon} & 43.6 & 95.2 & 93.6 & 90.6 & 94.5 & 94.7 \\
    
    \midrule
    \multicolumn{7}{c}{$with \ Self\text{-}Supervised \ Representation \ Learning$ (\textsc{Mlp-Linear})} \\
    \midrule 
    \textcolor{blue}{$\bullet$}Point-MAE$^\dagger$~\cite{pang2022PointMAE} & 22.1 & 82.8$\pm$0.3 & 83.2$\pm$0.2 & 74.1$\pm$0.2 & 90.2$\pm$0.1 & 90.7$\pm$0.1 \\  
    \textcolor{blue}{$\bullet$}\textbf{Point-PQAE} & 22.1 & \textbf{89.3$\pm$0.3} & \textbf{90.2$\pm$0.4} & \textbf{80.8$\pm$0.1} & \textbf{92.0$\pm$0.2} & {\textbf{92.2$\pm$0.1}} \\
    
    \midrule
    \multicolumn{7}{l}{Methods using cross-modal information and teacher models} \\
    \textcolor{blue}{$\bullet$}ACT~\cite{dong2022ACT} & 22.1 & 85.2$\pm$0.8 & 85.8$\pm$0.2 & 76.3$\pm$0.3 & 91.4$\pm$0.2 & 91.8$\pm$0.2 \\
    \textcolor{blue}{$\bullet$}ReCon~\cite{qi2023Recon} & 43.6 & 89.5$\pm$0.2 & 89.7$\pm$0.2 & 81.4$\pm$0.1 & 92.5$\pm$0.2 & 92.7$\pm$0.1 \\

    \midrule 
    \multicolumn{7}{c}{$with \ Self\text{-}Supervised \ Representation \ Learning$ (\textsc{Mlp-3})} \\
    \midrule 
    \textcolor{blue}{$\bullet$}Point-MAE$^\dagger$~\cite{pang2022PointMAE} & 22.1 & 85.8$\pm$0.3 & 85.5$\pm$0.2 & 80.4$\pm$0.2 & 91.3$\pm$0.2 & 91.7$\pm$0.2 \\ 
    \textcolor{blue}{$\bullet$}\textbf{Point-PQAE} & {22.1} & \textbf{90.7$\pm$0.2} & \textbf{90.9$\pm$0.2} & \textbf{83.3$\pm$0.1} & \textbf{92.8$\pm$0.1} & \textbf{92.9$\pm$0.1} \\
    \midrule
    \multicolumn{7}{l}{Methods using cross-modal information and teacher models} \\
    \textcolor{blue}{$\bullet$}ACT~\cite{dong2022ACT} & 22.1 & 87.1$\pm$0.2 & 87.9$\pm$0.4 & 81.5$\pm$0.2 & 92.7$\pm$0.2 & 93.0$\pm$0.1 \\
    \textcolor{blue}{$\bullet$}ReCon~\cite{qi2023Recon} & 43.6 & 90.6$\pm$0.2 & 90.7$\pm$0.3 & 83.8$\pm$0.4 & 93.0$\pm$0.1 & 93.4$\pm$0.1 \\
    
    \bottomrule[1pt]
    \end{tabular}
    \label{tab:scanobjnet-modelnet-cls}
    }
    \vspace{-9pt}
\end{table*}

{\bf 3D clean object classification.} 
We evaluate the performance of our pre-trained model for object classification using the ModelNet40 dataset~\cite{wu2015ModelNet}.  ModelNet40 comprises 12,311 meticulously crafted 3D CAD models, representing 40 distinct object categories. During testing, we adhere to the standard voting method~\cite{liu2019voting} to ensure fair comparisons with previous work such as~\cite{pang2022PointMAE,chen2024pointGPT}. Standard random scaling and random translation are applied for data augmentation during training, and the input point clouds exclusively contain coordinate information, without additional normal information provided. The results are presented in \cref{tab:scanobjnet-modelnet-cls}. Our Point-PQAE yields better or comparable results compared to previous approaches.


{\bf Few-shot learning.} 
We conduct the few-shot learning classification on the ModelNet40 dataset to show the generalization ability of our method following the protocols in previous studies~\cite{pang2022PointMAE,yu2022pointBert,sharma2020fewshotLearning,wang2021occlusionCompletion}. The few-shot learning experiments are conducted in the form of ``$w$-way, $s$-shot" including four parts where $w \in \{5,10\}$ represents the number of randomly selected classes and $s \in \{10,20\}$ means the number of randomly selected samples for each $w$. Each part is conducted with 10 independent trials. The mean accuracy and standard deviation are reported in \cref{tab:few-shot}. Our Point-PQAE demonstrates superior performance compared to previous methods in few-shot learning, even surpassing cross-modal methods that utilize strong pre-trained teachers such as ACT and ReCon under \textsc{Full} protocol. Specifically, Point-PQAE achieves 2.7\%, 1.1\%, 4.5\%, 1.9\% improvement under \textsc{Mlp-Linear} protocol and shows enhanced performance over the self-construction method Point-MAE.

\begin{table}[tb!]
    \centering
    \caption{Few-shot classification results on ModelNet40. 10 independent trials are conducted in each experimental setting. The mean accuracy (\%) and standard deviation are reported for each setting. The dagger($^\dagger$) denotes the baseline results reported from ReCon~\cite{qi2023Recon} which aligns augmentation with us and ReCon.}
    \vspace{-5pt}
    \resizebox{0.9\linewidth}{!}{
    \begin{tabular}{lcccc}
    \toprule[1pt]
    \multicolumn{1}{c}{\multirow{2}{*}{Methods}} & \multicolumn{2}{c}{5-way} & \multicolumn{2}{c}{10-way} \\ 
     & 10-shot & 20-shot & 10-shot & 20-shot \\ 
    \midrule
    \textcolor{orange}{$\circ$}DGCNN~\cite{wang2019DGCNN} & 31.6$\pm$2.8 & 40.8$\pm$4.6 & 19.9$\pm$2.1 & 16.9$\pm$1.5 \\
    \textcolor{orange}{$\circ$}OcCo~\cite{wang2021OcCo} & 90.6$\pm$2.8 & 92.5$\pm$1.9 & 82.9$\pm$1.3 & 86.5$\pm$2.2 \\
    \midrule
    \multicolumn{5}{c}{$with \ Self\text{-}Supervised \ Representation \ Learning$ (\textsc{Full})} \\
    \midrule 
    \textcolor{blue}{$\bullet$}Transformer~\cite{vaswani2017attentionIsallyouneed} & 87.8$\pm$5.2 & 93.3$\pm$4.3 & 84.6$\pm$5.5 & 89.4$\pm$6.3 \\
    \textcolor{blue}{$\bullet$}Point-BERT~\cite{yu2022pointBert} & 94.6$\pm$3.1 & 96.3$\pm$2.7 & 91.0$\pm$5.4 & 92.7$\pm$5.1 \\
    \textcolor{blue}{$\bullet$}MaskPoint~\cite{liu2022maskedpoint} & 95.0$\pm$3.7 & 97.2$\pm$1.7 & 91.4$\pm$4.0 & 93.4$\pm$3.5 \\
    \textcolor{blue}{$\bullet$}Point-MAE~\cite{pang2022PointMAE} & 96.3$\pm$2.5 & 97.8$\pm$1.8 & 92.6$\pm$4.1 & 95.0$\pm$3.0 \\
    \textcolor{blue}{$\circ$}Point-M2AE~\cite{zhang2022pointM2AE} & 96.8$\pm$1.8 & 98.3$\pm$1.4 & 92.3$\pm$4.5 & 95.0$\pm$3.0 \\ 
    \textcolor{blue}{$\bullet$}Point-MAE$^\dagger$~\cite{zhang2022pointM2AE} & 96.4$\pm$2.8 & 97.8$\pm$2.0 & 92.5$\pm$4.4 & 95.2$\pm$3.9 \\ 
    
    \textcolor{blue}{$\bullet$}\textbf{Point-PQAE} & {\textbf{96.9$\pm$3.2}} & \textbf{98.9$\pm$1.0} & \textbf{94.1$\pm$4.2} & \textbf{96.3$\pm$2.7} \\
    \midrule
    \multicolumn{5}{l}{Methods using cross-modal information and teacher models} \\
    \textcolor{blue}{$\bullet$}Joint-MAE~\cite{guo2023jointMAE} & 96.7$\pm$2.2 & 97.9$\pm$1.9 & 92.6$\pm$3.7  & 95.1$\pm$2.6 \\
    \textcolor{blue}{$\bullet$}TAP~\cite{wang2023TAP} & 97.3$\pm$1.8 & 97.8$\pm$1.9 & 93.1$\pm$2.6  & 95.8$\pm$1.0 \\
    
    \textcolor{blue}{$\bullet$}ACT~\cite{dong2022ACT} & 96.8$\pm$2.3 & 98.0$\pm$1.4 & 93.3$\pm$4.0  & 95.6$\pm$2.8 \\
    \textcolor{blue}{$\circ$}I2P-MAE~\cite{zhang2023I2PMAE} & 97.0$\pm$1.8 & 98.3$\pm$1.3 & 92.6$\pm$5.0  & 95.5$\pm$3.0 \\
    \textcolor{blue}{$\bullet$}ReCon~\cite{qi2023Recon} & 97.3$\pm$1.9 & 98.9$\pm$1.2 & 93.3$\pm$3.9 & 95.8$\pm$3.0 \\
    
    \midrule
    \multicolumn{5}{c}{$with \ Self\text{-}Supervised \ Representation \ Learning$ (\textsc{Mlp-Linear})} \\
    \midrule
    \textcolor{blue}{$\bullet$}Point-MAE$^\dagger$~\cite{zhang2022pointM2AE} & 91.1$\pm$5.6 & 91.7$\pm$4.0 & 83.5$\pm$6.1 & 89.7$\pm$4.1 \\ 
    \textcolor{blue}{$\bullet$}\textbf{Point-PQAE} & \textbf{93.0$\pm$4.6} & \textbf{96.8$\pm$1.9} & \textbf{89.0$\pm$5.2} & \textbf{93.5$\pm$3.8} \\
    
    
    \midrule
    \multicolumn{5}{c}{$with \ Self\text{-}Supervised \ Representation \ Learning$ (\textsc{Mlp-3})} \\
    \midrule
    \textcolor{blue}{$\bullet$}Point-MAE$^\dagger$~\cite{zhang2022pointM2AE} & 95.0$\pm$2.8 & 96.7$\pm$2.4 & 90.6$\pm$4.7 & 93.8$\pm$5.0 \\ 
    \textcolor{blue}{$\bullet$}\textbf{Point-PQAE} & {\textbf{95.3$\pm$3.4}} & \textbf{98.2$\pm$1.8} & \textbf{92.0$\pm$3.8} & \textbf{94.7$\pm$3.5}\\
    \bottomrule[1pt]
    \end{tabular}
     }
    \label{tab:few-shot}
    \vspace{-10pt}
\end{table}

{\bf Part segmentation.} 
We conduct part segmentation experiments on the ShapeNetPart~\cite{yi2016ShapNetPart} to validate the effectiveness of our Point-PQAE. The ShapeNetPart contains 16,881 objects covering 16 categories. We sample 2,048 points from each input point cloud, following previous work~\cite{yu2022pointBert,pang2022PointMAE} and divide each cloud into 128 point patches We use the same segmentation head and utilize learned features from the 4th, 8th, and 12th layers of the Transformer block as in Point-MAE~\cite{pang2022PointMAE}. We concatenate three levels of features. Then average pooling, max pooling, and upsampling are utilized to generate features for each point and an MLP is applied for label prediction. The results are reported in~\cref{tab:part_seg_shapenet}, The results are reported in~\cref{tab:part_seg_shapenet}, which show that Point-PQAE achieves a comparable Inst.mIoU to the previous method, achieving 84.6\% in Cls.mIoU and improving the from-scratch baseline by 1.4\%.

{\bf 3D scene segmentation.} 
Semantic segmentation on large-scale 3D scenes is challenging, requiring models possessing a deep comprehension of semantics and intricate local geometric relationships. We report semantic segmentation results on the S3DIS dataset~\cite{armeni2016S3DIS} in~\cref{tab:sem_seg_s3dis}. Our Point-PQAE improves the from-scratch baseline by 2.0\% and 1.4\%, and outperforms the self-reconstruction method Point-MAE by 0.7\% and 0.6\% in mAcc and mIoU.

\vspace{-4pt}
\subsection{Ablation study}
\vspace{-4pt}
Experiments are conducted to show the properties of our Point-PQAE. We report the classification results on three variants of ScanObjectNN. The experiment settings are aligned with \cref{real_world_cls}.

\begin{table}[tb!]
    \centering
    \caption{Segmentation results. Cls.mIoU (\%) and Inst.mIoU (\%) refer to Mean intersection over union for all classes and all instances, respectively. mAcc (\%) refers to mean accuracy.}
    \fontsize{25}{25}\selectfont
    \label{tab:part_seg}
    \vspace{-5pt}
    \begin{subtable}[t]{0.48\linewidth} 
        \centering
        \subcaption{Part segmentation on ShapeNetPart. }
        \vspace{5pt} 
        \resizebox{\linewidth}{!}{
        \begin{tabular}{lcc}
        \toprule[1.5pt]
        Method & Cls.mIoU (\%) & Inst.mIoU (\%) \\
        \midrule
        \multicolumn{3}{l}{\textit{Supervised}} \\
        PointNet~\cite{qi2017pointnet} & 80.4 & 83.7 \\
        PointNet++~\cite{qi2017pointnet++} & 81.9 & 85.1 \\
        DGCNN~\cite{wang2019DGCNN} & 82.3 & 85.2 \\
        \midrule
        \multicolumn{3}{l}{\textit{Self-supervised}} \\ 
        Transformer~\cite{vaswani2017attentionIsallyouneed} & 83.4 & 84.7 \\
        CrossPoint~\cite{afham2022crosspoint} & - & 85.5 \\
        Point-BERT~\cite{yu2022pointBert} & 84.1 & 85.6 \\
        Point-MAE~\cite{pang2022PointMAE} & 84.2 & \textbf{86.1} \\
        \textbf{Point-PQAE} (Ours) & \textbf{84.6} & \textbf{86.1} \\
        \midrule
        \multicolumn{3}{l}{\textit{Cross-modal}} \\ 
        ACT~\cite{dong2022ACT} & 84.7 & 86.1 \\
        ReCon~\cite{qi2023Recon} & 84.8 & {86.4} \\
        \bottomrule[1.5pt]
        \end{tabular}}
        \label{tab:part_seg_shapenet}
    \end{subtable}
    \hfill 
    \begin{subtable}[t]{0.48\linewidth} 
        \centering
        \subcaption{Semantic segmentation results on S3DIS Area 5.}
        \vspace{5pt} 
        \resizebox{\linewidth}{!}{
        \begin{tabular}{lcc}
        \toprule[1.5pt]
        Method & mAcc (\%) & mIoU (\%) \\
        \midrule
        \multicolumn{3}{l}{\textit{Supervised}} \\
        PointNet~\cite{qi2017pointnet} & 49.0 & 41.1 \\
        PointNet++~\cite{qi2017pointnet++} & 67.1 & 53.5 \\
        \midrule
        \multicolumn{3}{l}{\textit{Self-supervised}} \\
        Transformer~\cite{vaswani2017attentionIsallyouneed} & 68.6 & 60.0 \\
        Point-MAE~\cite{pang2022PointMAE} & 69.9 & 60.8 \\
        \textbf{Point-PQAE} (Ours) & \textbf{70.6} & \textbf{61.4} \\
        \midrule
        \multicolumn{3}{l}{\textit{Cross-modal}} \\ 
        ACT~\cite{dong2022ACT} & 71.1 & 61.2 \\
        \bottomrule[1.5pt]
        \end{tabular}}
        \label{tab:sem_seg_s3dis}
    \end{subtable}
    \vspace{-15pt}
\end{table}


{\bf View-relative positional embedding.} 
\label{sec:abla_embedding}
To analyze the effect of view-relative positional embedding in the positional query block, we conduct a series of experiments using different types of positional embeddings. We primarily consider two types of embeddings: Sinusoid (as discussed in ~\cref{sec:pq}) and Learnable, along with a None group. Learnable embedding refers to utilizing the relative position information $\mathbf{RP}_{1\rightarrow2}$ and $\mathbf{RP}_{2\rightarrow1}$ as input and employing an MLP composed of two linear layers with an activation function to map the 6-dimensional input to a D-dimensional output. The None group involves using randomly assigned embeddings to assess the effectiveness of VRPE design.
\cref{tab:sin-cos-abla} displays the results of different positional embeddings. In the None group, the model cannot learn the relative positional information across two views, leading to a significant drop in accuracy. 
In addition to the type of VRPE, we also experiment with Absolute Positional Embedding (APE) to compare its performance with our proposed VRPE. The APE is unsuitable for our framework since it requires a shared coordinate system for both views, making it incompatible with the normalization and rotation used in our method. We incorporate APE into our Point-PQAE by removing these operations. The results in \cref{tab:sin-cos-abla} show APE performs worse and support our claim.


\begin{table}[tb!]
    \centering
    \caption{Ablation study with Point-PQAE pre-training on  ShapeNet. The classification results by accuracy (\%) on three variants of ScanObjectNN are reported. Default settings are marked in \colorbox{gray!50}{gray}.}
    \vspace{-10pt}
    \begin{subtable}{0.35\textwidth}
        \centering
        \caption{View-Relative positional embedding}
        \resizebox{\linewidth}{!}{
        \begin{tabular}{cccc}
            \toprule[0.8pt]
            Positional Embedding & OBJ\underline{~~}BG & OBJ\underline{~~}ONLY & PB\underline{~~}T50\underline{~~}RS \\
            \midrule
            None & 84.5 & 85.9 & 79.3 \\
            APE (sinusoid) &  92.3 & 91.0 & 87.7  \\
            VRPE (learnable) & 93.4 & 93.1  & 89.1 \\
            \rowcolor{gray!50} VRPE (sinusoid~\cref{eq:sin-cos-1}) & {95.0} & {93.6} & {89.6} \\
            \bottomrule[0.8pt]
        \end{tabular}}
        \label{tab:sin-cos-abla}
    \end{subtable}
    \begin{subtable}{0.35\textwidth}
        \centering
        \caption{Data augmentation}
        \resizebox{\textwidth}{!}{
        \begin{tabular}{cccc}
        \toprule[0.8pt]
             Data augmentation & OBJ\underline{~~}BG & OBJ\underline{~~}ONLY & PB\underline{~~}T50\underline{~~}RS \\
        \midrule
             jitter & 93.3 & 91.4 & 87.3 \\
             scale & 93.3 & 91.7 & 88.0 \\
             \rowcolor{gray!50} {rotation} & {95.0} & {93.6} & {89.6} \\
             scale\&translate & 92.8 & 91.7 & 88.1 \\
             rotation+scale\&translate & 93.8 & 92.9 & 89.1 \\
        \bottomrule[0.8pt]
        \end{tabular}}
        \label{tab:data-aug}    
    \end{subtable}
    \begin{subtable}{0.35\textwidth}
        \centering
        \caption{3D random crop mechanism}
        \resizebox{\textwidth}{!}{
        \begin{tabular}{ccccc}
        \toprule[0.8pt]
             Method & crop & OBJ\underline{~~}BG & OBJ\underline{~~}ONLY & PB\underline{~~}T50\underline{~~}RS \\
        \midrule
             Point-MAE & \wrongSymbol & 92.6 & 91.9 & 88.4 \\
             Point-MAE & \rightSymbol & 92.9 & 92.1 &  88.8 \\
             Point-PQAE & \wrongSymbol & 92.9 & 92.3 & 87.7 \\
             \rowcolor{gray!50} Point-PQAE & \rightSymbol & 95.0 & 93.6 & 89.6 \\
        \bottomrule[0.8pt]
        \end{tabular}}
        \label{tab:ablation_on_crop}
    \end{subtable}
    \begin{subtable}{0.35\textwidth}
        \caption{Effects of augmentations after random crop}
        \resizebox{\linewidth}{!}{
        \begin{tabular}{ccccc}
        \toprule
        norm. & rotation & OBJ\underline{~~}BG & OBJ\underline{~~}ONLY & PB\underline{~~}T50\underline{~~}RS \\
        \midrule
        \wrongSymbol & \wrongSymbol & 92.8 & 92.1 & 88.2 \\
        \rightSymbol & \wrongSymbol & 93.3 & 92.6 & 87.9 \\
        \wrongSymbol & \rightSymbol & 93.5 & 92.6 & 88.8 \\
        \rowcolor{gray!50} \rightSymbol & \rightSymbol & 95.0 & 93.6 & 89.6 \\
        \bottomrule
        \end{tabular}
        }
        \label{tab:aug_after_crop}
    \end{subtable}
    \vspace{-15pt}
\end{table}

{\bf Data augmentation.} 
After applying the random crop mechanism to the point cloud, we further perform data augmentation on the cropped views to generate new views. Data augmentation is crucial for generating diverse and decoupled views, so we conducted ablation studies on different augmentation methods. The results in \cref{tab:data-aug} indicate that rotation performs best.

{\bf Effectiveness of the crop mechanism.}
Random crop has been shown to be highly effective in 2D self-supervised learning (SSL), particularly for contrastive learning~\cite{grill2020BYOL, chen2020SimCLR, bao2021beit}, and is also vital for generative learning~\cite{he2022VisionMAE}. However, its potential in 3D SSL remains largely unexplored. Even contrastive learning methods for point cloud understanding have not explored the power of it~\cite{afham2022crosspoint, pang2023crop_related}. To address this gap, we first introduce a custom-designed random crop mechanism for point cloud data, using it as the foundation for generating decoupled views. Without the crop augmentation, the two views will contain exactly the same shape though with normalization and rotation applied. \textbf{The model would only need to infer the augmentation rather than the inter-view relationship, which means the views are not decoupled.} 
Conversely, given two isolated cropped point clouds, followed by min-max normalization and rotation for further decoupling, there remain overlapping and non-overlapping parts, requiring the model to effectively encode the intra-view parts in order to infer the unknown inter-view points in our Point-PQAE. However, directly applying random crop as an additional augmentation to the self-reconstruction method (Point-MAE) does not improve the performance much, as shown in \cref{tab:ablation_on_crop}. In other words, \textbf{it means our Point-PQAE fits crop mechanism better by harmoniously incorporating it into the decoupled views generation process.} The experiment results in \cref{tab:ablation_on_crop} demonstrate the critical role of random crop in our method. We set minimum crop ratio $r_m=1.0$ when removing the crop in our Point-PQAE. 

{\bf Analysis on decoupled views generation.} While crop plays an important role in our method, relying solely on the random crop mechanism to generate new views would degenerate cross-reconstruction to simply reconstructing one part from another. This is similar to the block mask self-reconstruction used in Point-MAE~\cite{pang2022PointMAE}, where visible blocks are used to reconstruct masked blocks, which is also trivial for pre-training. The only difference between them is that the cropped parts can overlap somewhat, whereas the block mask does not. What makes the cross-reconstruction different from block mask self-reconstruction are the subsequent augmentations including normalization and rotation that decouple the cropped parts. By normalizing cropped point clouds centered on the geometric centers, the coordinate systems of the two point clouds become isolated and independent from each other, with rotation further amplifying the variance between the two views, thus enabling the generation of two decoupled views. 
The importance of the augmentations applied after the random crop can be validated through experiments, as shown in \cref{tab:aug_after_crop}.

\vspace{-6pt}
\section{Conclusion}
\vspace{-4pt}
In this paper, we propose a novel cross-reconstruction generative framework for self-supervised learning on 3D point clouds, called Point-PQAE. In contrast to well-studied self-reconstruction schemes, Point-PQAE reconstructs one cropped point cloud from another decoupled point cloud.  To supply sufficient information for cross-reconstruction, we further propose a 3D view-relative positional embedding and a corresponding position-aware query module. Compared to self-reconstruction, the cross-reconstruction task, carefully designed by us, is much more challenging for pre-training, promoting the learning of richer semantic representations during pre-training. This enables our proposed Point-PQAE to outperform previous single-modal self-reconstruction methods by a margin and to perform on par with, or better than cross-modal methods.

\clearpage

\appendix

\section*{Appendix}

\section{Comparisons to more peer methods}
Apart from the self-reconstruction methods in the main paper (including baseline Point-MAE~\cite{pang2022PointMAE} and others), there are some other peer methods, including Point-FEMAE~\cite{zha2023PointFEMAE}, PCP-MAE~\cite{zhang2024pcpmae}, I2P-MAE~\cite{zhang2023I2PMAE}, Joint-MAE~\cite{guo2023jointMAE}, Cross-BERT~\cite{li2023crossBert}, and TAP~\cite{wang2023TAP}. Here, we discuss the relation of Point-PQAE with these approaches and compare its performance against them to better position our work. A brief comparison of peer methods with our Point-PQAE can be seen in~\cref{tab:brief_comparison}, and the performance of these methods on downstream tasks is reported in~\cref{tab:supp_compare}. Point-PQAE achieves the best or comparable performance when compared with them.

\textbf{Relation to Point-FEMAE~\cite{zha2023PointFEMAE}.} \textbf{Connection.} Both of them are reconstruction-based methods. \textbf{Differences.} \textbf{1) }{\it Pre-Training Efficiency.} Point-FEMAE performs mask reconstruction in both the global and local branches and introduces Local Enhancement Module (LEM) which consists of some convolution layers and MLP layers to each transformer block. To achieve local patch convolution with coordinate-based nearest neighbors, when tokens are input to LEM, it duplicates K nearest neighboring patches ($k=20$) for each input token and aggregates nearest information for each token which brings an extra calculation burden to each block in the encoder. Point-PQAE utilizes original transformer blocks, making it more efficient during pre-training. \textbf{2) }{\it Backbone.} Point-FEMAE reserves the LEM modules in the encoder for fine-tuning which means adding convolution and MLP layers for each transformer block to the backbone for fine-tuning, while our Point-PQAE utilizes an encoder consisting of pure transformer blocks for fine-tuning, which remains simple and is aligned to previous work.


\textbf{Relation to PCP-MAE~\cite{zhang2024pcpmae}.} \textbf{Connection.} Both of them are reconstruction-based methods. \textbf{Differences.} Targeted at alleviating information leakage of centers in point cloud, PCP-MAE proposes a new module called Predicting Center Module (PCM) and a novel loss for better utility of centers based on Point-MAE, which is still a self-reconstruction method. Our Point-PQAE differs from it as it is a pioneering cross-reconstruction method; it uses VRPE to perform cross-view point cloud reconstruction, which overcomes the limitations of self-reconstruction methods.

\textbf{Relation to I2P-MAE~\cite{zhang2023I2PMAE}.} \textbf{Connection.} Both of them are reconstruction-based methods. \textbf{Differences.} I2P-MAE heavily relies on strong pre-trained 2D models as a guide to achieve multi-task cross-modal learning while our Point-PQAE utilizes single-modal data without relying on any pre-trained model. Besides, the utilization of 2D data in I2P-MAE brings a heavy computation burden to the pre-training process.

\begin{table*}[tb!]
    \centering
    \caption{Methodology comparisons between our Point-PQAE and other peer methods. The ``Extra'' in the table means extra parts in contrast to Point-MAE which adopts standard transformer blocks as backbone.}
    \resizebox{.8\textwidth}{!}{
    \begin{tabular}{lcccccc}
    \toprule[1pt]
        \multirow{2}{*}{Methods} & Single-/Cross- & Pre-trained  & Single-/Multi- & Extra Transformer & Extra Modules \\
        & Modal & Model Needed & Task & Blocks (Pre-training) & (Fine-tuning) \\
    \midrule
    Point-MAE~\cite{pang2022PointMAE} & {Single} & {\usym{2717}} & {Single} & {\usym{2717}} & {\usym{2717}} \\
    Point-FEMAE~\cite{zha2023PointFEMAE} & Single & \usym{2717} & Multi & \usym{2717} & \usym{2713} \\
    PCP-MAE~\cite{zhang2024pcpmae} & Single & \usym{2717} & Multi & \usym{2717} & \usym{2717} \\
    I2P-MAE~\cite{zhang2023I2PMAE} & Cross & \usym{2713} & Multi & \usym{2717} & \usym{2717} \\
    Joint-MAE~\cite{guo2023jointMAE} & Cross & \usym{2717} & Multi & \usym{2713} & \usym{2717} \\
    Cross-BERT~\cite{li2023crossBert} & Cross & \usym{2713} & Multi & \usym{2713} & \usym{2717} \\
    TAP~\cite{wang2023TAP} & Cross & \usym{2717} & Single & \usym{2717} & \usym{2717} \\
    \midrule
    Point-PQAE & Single & \usym{2717} & Single & \usym{2717} & \usym{2717} \\
    \bottomrule[1pt]
    \end{tabular}}
    \label{tab:brief_comparison}
\end{table*}

\begin{table*}[tb!]
    \centering
    \caption{Performance of peer methods. The classification results on ScanObjectNN and ModelNet40 and few-shot learning results on ModelNet40 are reported by accuracy (\%). We term OBJ\underline{~~}BG, OBJ\underline{~~}ONLY, PB\underline{~~}T50\underline{~~}RS as BG, OY, RS respectively. We compare methods using the \textcolor{blue}{$\bullet$} plain Transformer architectures, \eg Point-MAE\cite{pang2022PointMAE}, Point-PQAE (ours), \textcolor{blue}{$\circ$} hierarchical Transformer architectures and \textcolor{purple}{$\circ$} methods with extra modules during fine-tuning.}
    \resizebox{.9\textwidth}{!}{
    \begin{tabular}{lcccccccccc}
    \toprule[1pt]
    \multirow{3}{*}{Methods} & \multirow{3}{*}{\#P} & \multicolumn{3}{c}{ScanObjectNN} & \multicolumn{2}{c}{ModelNet40} & \multicolumn{4}{c}{ModelNet40 few-shot} \\
     & & \multirow{2}{*}{BG} & \multirow{2}{*}{OY} & \multirow{2}{*}{RS} & \multirow{2}{*}{1K P} & \multirow{2}{*}{8K P} & \multicolumn{2}{c}{5-way} & \multicolumn{2}{c}{10-way} \\
     & & & & & & & 10-shot & 20-shot & 10-shot & 20-shot \\
    \midrule 
    \textcolor{blue}{$\bullet$}Point-MAE~\cite{pang2022PointMAE} & 22.1 & 90.0 & 88.3 & 85.2 & 93.8 & 94.0 & 96.3$\pm$2.5 & 97.8$\pm$1.8 & 92.6$\pm$4.1 & 95.0$\pm$3.0 \\
    \textcolor{purple}{$\circ$}Point-FEMAE~\cite{zha2023PointFEMAE} & 27.4 & \underline{95.2} & 93.3 & \underline{90.2} & \textbf{94.5} & - & 97.2$\pm$1.9 & 98.6$\pm$1.3 & \textbf{94.0$\pm$3.3} & 95.8$\pm$2.8 \\ 
    \textcolor{blue}{$\bullet$}PCP-MAE~\cite{zhang2024pcpmae} & 22.1 & \textbf{95.5} & \textbf{94.3} & \textbf{90.4} & \underline{94.2} & - & \textbf{97.4$\pm$2.3} & \textbf{99.1$\pm$0.8} & \underline{93.5$\pm$3.7} & \underline{95.9$\pm$2.7} \\ 
    \textcolor{blue}{$\circ$}I2P-MAE~\cite{zhang2023I2PMAE} & - & 94.2 & 91.6 & 90.1 & 94.1 & - & 97.0$\pm$1.8 & 98.3$\pm$1.3 & 92.6$\pm$5.0 & 95.5$\pm$3.0 \\ 
    \textcolor{purple}{$\circ$}Joint-MAE~\cite{guo2023jointMAE} & - & 90.9 & 88.9 & 86.1 & 94.0 & - & 96.7$\pm$2.2 & 97.7$\pm$1.8 & 92.6$\pm$3.7 & 95.1$\pm$2.6 \\  
    \textcolor{blue}{$\bullet$}{Cross-BERT}~\cite{li2023crossBert} & 22.1 & 93.7 & 92.1 & 89.0 & \underline{94.2} & \textbf{94.4} & 97.0$\pm$2.1 & 98.2$\pm$1.3 & 93.0$\pm$3.4 & 95.6$\pm$3.0 \\
    \textcolor{blue}{$\bullet$}{TAP}~\cite{wang2023TAP} & 22.1 & 90.4 & 89.5 & 85.7 & - & - & \underline{97.3$\pm$1.8} & 97.8$\pm$1.7 & 93.1$\pm$2.6 & 95.8$\pm$1.0 \\
    \midrule
    \textcolor{blue}{$\bullet$}{Point-PQAE} & {22.1} & {95.0} & \underline{93.6} & {89.6} & {93.9} & \underline{{94.3}} & 96.9$\pm$3.0 & \underline{99.0$\pm$1.0} & \textbf{94.0$\pm$4.0} & \textbf{96.1$\pm$2.8} \\ 
    \bottomrule[1pt]
    \end{tabular}
    \label{tab:supp_compare}
    }
\end{table*}

\textbf{Relation to Joint-MAE~\cite{guo2023jointMAE}.} \textbf{Connection.} Both of them are reconstruction-based methods. \textbf{Differences.} Joint-MAE utilizes a shared weight encoder but 3 different decoders for pre-training, and it's a multi-task pre-training method including 2D / 3D / 2D-3D reconstruction, which makes it more computational while Point-PQAE utilizes one encoder and one decoder only for per-training and is a single task method which focuses on 3D data cross-reconstruction. 

\textbf{Relation to Cross-BERT~\cite{li2023crossBert}.} \textbf{Connection.} Both are point cloud self-supervised methods. \textbf{Differences.} Cross-BERT is a method that utilizes two isolated encoders for cross-modal learning of point clouds and rendered images. To prevent the collapse of its intra-/cross- modal contrastive learning, it further uses another two momentum encoders that perform EMA-update, which makes pre-training of Cross-BERT much more complex than Point-PQAE. Additionally, Cross-BERT requires pre-training a dVAE as the tokenizer and includes a mask cross-modal learning task alongside contrastive learning. In contrast, Point-PQAE focuses solely on point clouds, utilizing a single encoder for single-task self-supervised pre-training.

\textbf{Relation to TAP~\cite{wang2023TAP}.} \textbf{Connection.} Both are reconstruction-based methods. \textbf{Differences.} TAP makes cross-modal reconstruction, which renders images of point clouds with different poses and after getting the latent representation of the 3D point cloud, it uses the pose information of the rendered images to query cross-modal information from the latent representation by cross-attention. And then using the queried information to rebuild the rendered image. Point-PQAE uses view-relative position embedding (VRPE) to make cross-view information interaction by cross-attention to achieve cross-reconstruction which uses 3D data only. Extra online operation on rendering and processing 2D data in TAP will raise the computational needs compared to Point-PQAE. 

{\section{{Related cross-reconstruction works}}}

Our Point-PQAE pioneers the cross-reconstruction paradigm in 3D point cloud self-supervised learning (SSL). \textbf{There are two essential components in our proposed cross-reconstruction framework:} 1) Two isolated/decoupled views, rather than two parts of the same instance that maintain a fixed relative relationship. 2) A model that achieves cross-view reconstruction using relative position information. To our knowledge, there are no similar previous methods in this domain. To better position our methodology, we compare it to similar SSL methods, including SiamMAE~\cite{gupta2023SiamMAE} and CropMAE~\cite{eymael2025CropMAE}, proposed in the image domain. SiamMAE operates on pairs of randomly sampled video frames and asymmetrically masks them, utilizing the past frame to predict the masked future frame. CropMAE relies on image augmentations to generate two views, using one to reconstruct the other.

{\bf Relation of SiamMAE and CropMAE to our Point-PQAE:} All are cross-reconstruction methods. Both SiamMAE and CropMAE have two essential components for cross-reconstruction-framework including two-view (different frames sampled from one video for SiamMAE and isolated augmented images for CropMAE) and cross-view reconstruction model. They can be treated as cross-reconstruction methods, similar to our Point-PQAE.

{\bf Difference of SiamMAE and CropMAE to our Point-PQAE:} \textbf{1) Different domain:} SiamMAE and CropMAE focus on the 2D SSL domain. Our Point-PQAE is the first method for cross-reconstruction in the 3D SSL domain. \textbf{2) Asymmetric/symmetric reconstruction:} SiamMAE uses the past to predict the future, which is asymmetric. CropMAE performs asymmetric reconstruction and doesn't explore siamese cross-reconstruction. In contrast, our Point-PQAE is inherently symmetric, and the siamese loss brings a performance gain. \textbf{3) No relative information utilized:} SiamMAE and CropMAE do not incorporate relative information into training but rely on non-fully masking to guide the cross-reconstruction. The VRPE adopted by our Point-PQAE provides explicit guidance, making training more stable and improving explainability. \textbf{4) No tuned-needed mask ratio exists in our framework.} There is a hyperparameter mask ratio that needs to be tuned in both SiamMAE and CropMAE, but this is not the case in our Point-PQAE framework.

{\bf Relation and differences between Joint-MAE~\cite{guo2023jointMAE} and PiMAE~\cite{chen2023pimae}.} We discuss the differences between our framework and two seemingly similar methods in the point cloud domain: Joint-MAE~\cite{guo2023jointMAE} and PiMAE~\cite{chen2023pimae}. Joint-MAE and PiMAE adopt a similar strategy, utilizing paired point clouds and images to perform cross-modal masked autoencoding. Our framework, however, differs significantly from these two methods. 

The \textbf{relation} of these methods to our Point-PQAE is that all three are self-supervised approaches that focus on the point cloud domain. 

\textbf{Differences:} 

\textbf{1) Different motivations:} Joint-MAE and PiMAE aim to explore the semantic correlation between 2D and 3D data by performing 3D-2D interactions and achieving cross-modal self-reconstruction through cross-modal knowledge. In contrast, inspired by the success of two-view pre-training paradigms, we propose Point-PQAE, the first cross-reconstruction framework for point cloud self-supervised learning (SSL). 

\textbf{2) Different modalities:} Both Joint-MAE and PiMAE rely on paired image-point cloud data, making them cross-modal methods. Our Point-PQAE, on the other hand, only consumes unlabeled point cloud data, making it more easily extendable. Additionally, incorporating image data could increase computational requirements. 

\textbf{3) Joint-MAE and PiMAE cannot be called cross-reconstruction methods, unlike our Point-PQAE, because:}

\begin{itemize}
    \item Recall that cross-reconstruction methods require two components: decoupled views and a cross-reconstruction framework. In cross-reconstruction, decoupled views are obtained through independent augmentations, achieving significant diversity between views, and the cross-reconstruction framework relies on information from view 1 to mandatorily reconstruct view 2.
    \item The paired 3D and 2D views used by Joint-MAE and PiMAE cannot be considered isolated or decoupled views. Take PiMAE as an example: the image is merely a render from a specific camera pose of the point cloud. No augmentations can be applied to either of these views (as discussed in Section 4 of the PiMAE paper), so diversity between views cannot be achieved.
    \item Cross-view knowledge is used as auxiliary, not mandatory, in these two methods. If either the 3D or 2D data is removed, reconstruction can still be achieved, which turns into the case in MAE~\cite{he2022VisionMAE} or Point-MAE~\cite{pang2022PointMAE}. However, a cross-reconstruction framework should mandatorily rely on view 1 to reconstruct view 2, as in our Point-PQAE, SiamMAE~\cite{gupta2023SiamMAE}, and CropMAE~\cite{eymael2025CropMAE}. For instance, in Joint-MAE, 3D information is used as auxiliary for 2D MAE (or vice versa), and a cross-reconstruction loss (specifically, cross-modal reconstruction loss) is added to the 2D-3D output.
\end{itemize}

Thus, it is more appropriate to refer to these methods as cross-modality self-reconstruction methods.

\section{Discussion on the view-relative positional embedding and positional query}

{\bf Relative Positional Embedding (RPE) methods.} To better position the View-Relative Positional Embedding (VRPE) proposed by us for point cloud cross-view reconstruction, we discuss the difference between our VRPE and existing RPE methods. In the fields of Natural Language Processing (NLP) and 2D vision, RPE techniques have been widely adopted~\cite{wu2021rethinkingRPE, yang2019xlnetNLPRPE, raffel2020exploringNLPRPE}. For instance, Rotary Positional Embedding (RoPE)~\cite{su2024RoPE} is an emerging RPE technique gaining traction in the realm of large language models (LLMs). RoPE integrates rotational transformations to encode relative token positions, enabling more efficient extrapolation over unseen sequences. {iRPE~\cite{wu2021rethinkingRPE} first reviews existing relative position encoding methods, and then proposes new RPE methods dedicated to 2D images. The work~\cite{qu2021RPE2} investigates the potential problems in Shaw-RPE and
XL-RPE, which are the most representative and prevalent RPEs, and proposes two novel RPEs called LRHC-RPE and GCDF-RPE.} Generally, in NLP and 2D vision, RPE captures the relative distances or orientations between tokens or pixels to enhance the model's capacity to understand relationships between paired elements. This approach often leads to improved generalization, especially when handling out-of-distribution data.

In contrast, our proposed VRPE is designed with a view- or instance-based focus, rather than a token-based one. Rather than capturing relationships between individual tokens or pixels, our VRPE encodes the positional relationships between two decoupled views. Our approach is not aimed at improving extrapolation or generalization. Instead, it is tailored to model the geometric and contextual information between different views to facilitate accurate cross-view reconstruction. This shift in focus makes our VRPE fundamentally distinct from existing RPE methods, highlighting the importance of carefully distinguishing our approach from existing RPE techniques.



{\bf Related Positional Query (PQ) methods.}
The positional query is also used in 2D self-supervised learning (SSL) and AI-generated content (AIGC). PQCL~\cite{pqcl} pioneered the introduction of positional query, aiming to represent geometric relationships between multiple cropped views. PQDiff~\cite{pqdiff} advanced this concept by devising a contiguous relative positional query module, applying it to image outpainting to achieve arbitrary location and contiguous expansion factor outpainting. Positional query has also found applications in 2D segmentation tasks. For example, DFPQ~\cite{he2023DFPQ} generates positional queries dynamically by leveraging cross-attention scores from the previous decoder block and the positional encodings of the image features, which together enhance the effectiveness of semantic segmentation. Our method, however, distinguishes itself from these existing positional query approaches by focusing on the 3D world, which presents significantly greater complexity (one more dimension) and challenges compared to 2D image domains. 
By leveraging the obtained VRPE to query the target view from the source view, our PQ technique successfully achieves decoupled view reconstruction.

\section{Additional experimental details}
\label{sec:supp_add_exp_details}

{\bf Training details.} 
We utilize ShapeNet~\cite{chang2015ShapeNet} as our pre-training dataset, which comprises a curated collection of 3D CAD object models, featuring 51K unique models across 55 common categories. The pre-training process spans 300 epochs, employing a cosine learning rate schedule~\cite{loshchilov2016cosLrDecay} starting at 5e-4, with a warm-up period of 10 epochs. We use the AdamW optimizer~\cite{loshchilov2017AdamWOptimizer} and a batch size of 128. All experiments are conducted on a single GPU \ie, RTX 3090 (24GB). For further training details including pre-training and finetuning, refer to~\cref{tab:supp_training_details}. During the pre-training of our Point-PQAE on ShapeNet, we apply rotation to the input point cloud following ReCon~\cite{qi2023Recon}, followed by generating decoupled views from the augmented point cloud.

\begin{table*}[tb!]
\centering
\caption{Training details for pretraining and downstream fine-tuning.}
\resizebox{0.9\textwidth}{!}{
\begin{tabular}{@{}lccccc@{}}
\toprule[1pt]
Config               & \textbf{ShapeNet} & \textbf{ScanObjectNN} & \textbf{ModelNet} & \textbf{ShapeNetPart} & \textbf{S3DIS} \\ \midrule
optimizer            & AdamW             & AdamW                 & AdamW             & AdamW                 & AdamW \\
learning rate        & 5e-4              & 2e-5                  & 1e-5              & 2e-4                 & 2e-4 \\
weight decay         & 5e-2              & 5e-2                  & 5e-2              & 5e-2                 & 5e-2 \\
learning rate scheduler & cosine          & cosine                & cosine            & cosine              & cosine \\
training epochs      & 300               & 300                   & 300               & 300                  & 60 \\
warmup epochs        & 10                & 10                    & 10                & 10                   & 10 \\
batch size           & 128               & 32                    & 32                & 16                   & 32 \\
drop path rate       & 0.1               & 0.2                   & 0.2               & 0.1                  & 0.1 \\
number of points     & 1024              & 2048                  & 1024         & 2048        & 2048           \\
number of point patches & 64             & 128                   & 64            & 128                      & 128 \\
point patch size     & 32                & 32                    & 32                & 32                   & 32 \\
augmentation         & Rotation          & Rotation              & Scale\&Trans      & -                & -     \\
\midrule
GPU device           & RTX 3090     &  RTX 3090            & RTX 3090        & RTX 3090          & RTX 3090  \\ \bottomrule[1pt]
\end{tabular}}
\label{tab:supp_training_details}
\end{table*}


{\bf Finetuning evaluation protocol.} For classification tasks on ScanObjectNN and ModelNet40, as well as few-shot learning on ModelNet40, we adopt three evaluation protocols, following~\cite{dong2022ACT, qi2023Recon}, to assess both the transferability of learned representations (\textsc{Full}) and the quality of frozen features (\textsc{Mlp-Linear}, \textsc{Mlp-3}). The protocols are as follows:

\begin{enumerate}[label=(\alph*)]
    \item \textsc{Full}: Fine-tuning the pre-trained model by updating both the backbone and the classification head.
    \item \textsc{Mlp-Linear}: Fine-tuning by updating only the classification head, which consists of a single-layer linear MLP.
    \item \textsc{Mlp-3}: Fine-tuning by updating only the parameters of a three-layer non-linear MLP classification head (which is structured the same as in \textsc{Full}).
\end{enumerate}



\section{Additional ablation study}

{\bf Integrate Positional Query (PQ) scheme into knowledge distillation.} 
\label{distill}
The knowledge distillation~\cite{hinton2015knowledgeDistillation} typically involves inputting the same instance into both the student model and the frozen teacher model, then maximizing the mutual agreement between their outputs to distill knowledge from the teacher to the student. Our positional query block can be seamlessly integrated into knowledge distillation, allowing for cross-view distillation rather than being confined to distillation within the same view. 
For example, view 1 is fed to the student, view 2 is fed to the teacher, and a positional query block is added after the backbone to model relative relations and recover the latent representation of view 2. We conduct experiments on distilling the pre-trained model ReCon~\cite{qi2023Recon}, and the results are reported in \cref{tab:abla_distill}, indicating that our PQ scheme successfully learns knowledge from the ReCon teacher and performs much better than the baseline. It shows that the PQ scheme can be easily utilized as a plug-in tool for knowledge distillation.

\begin{table}[tb!]
    \centering
    \caption{Integrate PQ into distillation. Results on ScanobjectNN (\%) are reported.}
    \vspace{-4pt}\resizebox{0.4\textwidth}{!}{
    \begin{tabular}{cccc}
    \toprule[0.8pt]
         Type & OBJ\underline{~~}BG & OBJ\underline{~~}ONLY & PB\underline{~~}T50\underline{~~}RS \\
    \midrule
         Train from scratch & 83.0 & 84.0 & 79.1 \\
         PQ distillation & 93.5 & 91.9 & 88.5 \\
    \bottomrule[0.8pt]
    \end{tabular}}
    \label{tab:abla_distill}
    \vspace{-6pt}
\end{table}



\begin{table}[tb!]
    \centering
    \caption{Reconstruction loss function. The default setting is marked in gray.}
    \vspace{-6pt}\resizebox{0.45\textwidth}{!}{
    \begin{tabular}{cccc}
    \toprule[0.8pt]
         Loss Function & OBJ\underline{~~}BG & OBJ\underline{~~}ONLY & PB\underline{~~}T50\underline{~~}RS \\
    \midrule
         cos & 90.5 & 89.8 & 85.2 \\
         CD-$l1$ & 93.1 & 91.7 & 89.4 \\
         \cellcolor{gray!50}{CD-$l2$} & \cellcolor{gray!50}{95.0} & \cellcolor{gray!50}{93.6} & \cellcolor{gray!50}{89.6} \\
    \bottomrule[0.8pt]
    \end{tabular}}
    \label{tab:ablation_on_loss}
\end{table}

{\bf Reconstruction loss function.} 
\cref{tab:ablation_on_loss} shows the performance of Point-PQAE using different reconstruction loss functions: cosine similarity loss (cos), $l1$-form Chamfer distance~\cite{fan2017chamferDist} (CD-$l1$), and the $l2$-form Chamfer distance (CD-$l2$). The results show the CD-$l2$ is more suitable for Point-PQAE. 

\begin{table}[tb!]
    \centering
    \caption{Siamese loss. The default setting is marked in gray.}
    \vspace{-6pt}\resizebox{0.48\textwidth}{!}{
    \begin{tabular}{cccc}
        \toprule[0.8pt]
        Loss Function & OBJ\underline{~~}BG & OBJ\underline{~~}ONLY & PB\underline{~~}T50\underline{~~}RS \\
        \midrule
        $\mathcal{L}_{2\rightarrow 1}$ & 93.4 & 92.4 & 89.2 \\
        \rowcolor{gray!50} {$\mathcal{L}_{2\rightarrow 1}+\mathcal{L}_{1\rightarrow 2}$} & {95.0} & {93.6} & {89.6}  \\
        \bottomrule[0.8pt]
    \end{tabular}}
    \label{tab:siamese-loss}
\end{table}

{\bf Siamese loss function.} 
The generative pre-training task designed by us is naturally a siamese structure and we get the form of $\mathcal{L}_{cross}=\mathcal{L}_{2\rightarrow 1}+\mathcal{L}_{1\rightarrow 2}$ as stated in~\cref{sec:siamese-loss}. We analyze the benefit of the siamese loss function by doing an ablation study with loss functions $\mathcal{L}_{cross}=\mathcal{L}_{2\rightarrow 1}+\mathcal{L}_{1\rightarrow 2}$ or $\mathcal{L}_{2\rightarrow 1}$ only. The~\cref{tab:siamese-loss} presents the experiment results. It shows this siamese loss function contributes to the performance of our Point-PQAE and brings accuracy gain.

\begin{figure*}[tb!]
\centering
\begin{minipage}[t]{0.33\textwidth}
    \centering
    \includegraphics[width=0.99\textwidth]{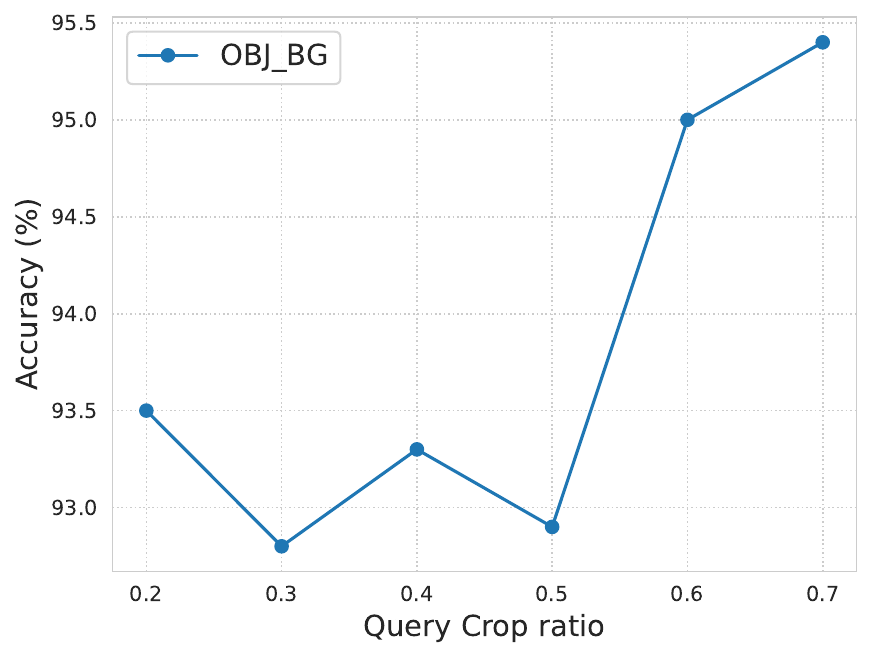}
    \label{fig:BG}
\end{minipage}%
\begin{minipage}[t]{0.33\textwidth}
    \centering
    \includegraphics[width=0.99\textwidth]{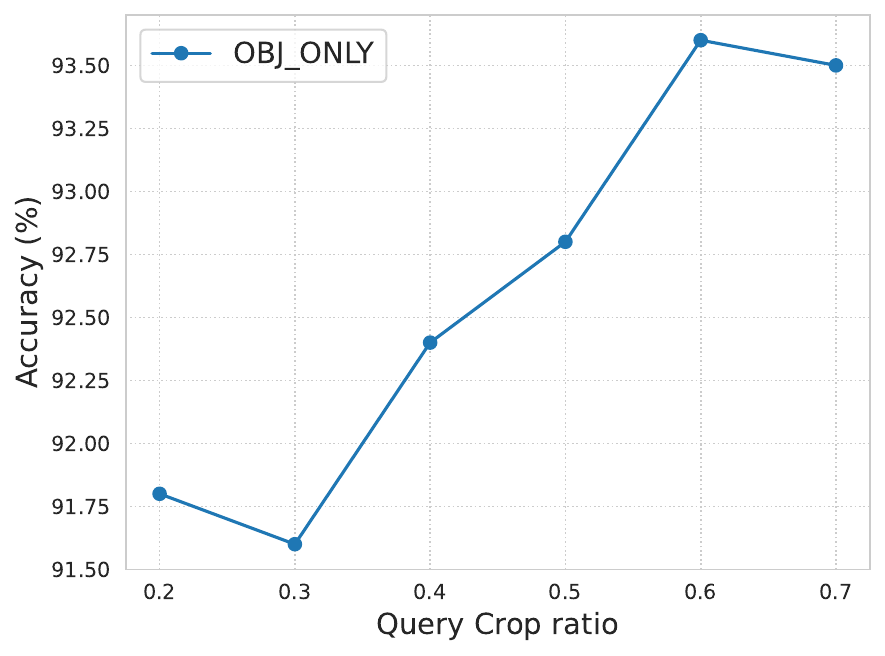}
    \label{fig:ONLY}
\end{minipage}%
\begin{minipage}[t]{0.33\textwidth}
    \centering
    \includegraphics[width=0.975\textwidth]{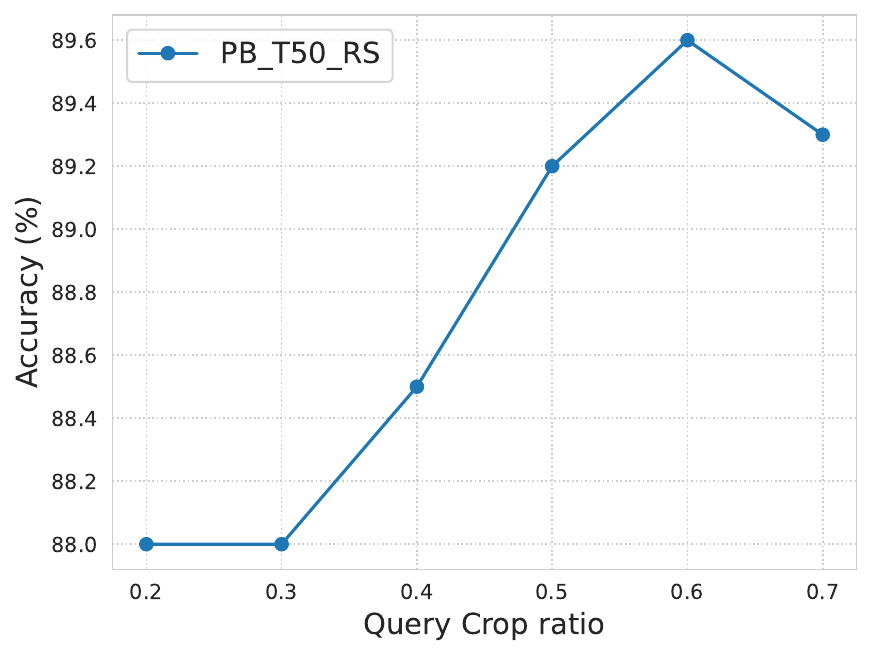}
    \label{fig:HARD}
\end{minipage}%
\vspace{-15pt}
\caption{Ablation study on different minimum crop ratios $r_m$, where the results (\%) of three variants: OBJ\underline{~~}BG, OBJ\underline{~~}ONLY, PT\underline{~~}T50\underline{~~}RS on ScanObjectNN are reported.}
\label{fig:ablation_crop_ratio}
\vspace{-7pt}
\end{figure*}

{\bf Minimum crop ratio.} 
The minimum crop ratio $r_m$ is important for the proposed point cloud crop mechanism. We conduct experiments to analyze the effect of minimum random crop ratios on the performance. {The results are reported in \cref{fig:ablation_crop_ratio}.} The results show that 0.6 is the best crop ratio for our Point-PQAE. When the ratio is too low, the model struggles to extract sufficient relevant information from the cropped view for effective cross-reconstruction. Conversely, excessively high ratios make the task too straightforward, hindering the model from learning robust representations.


{\bf Definition of views and parts in our work.} We emphasize the importance of distinguishing between parts and views to understand the significance of decoupled view generation and our cross-view reconstruction method. We define the following:

\begin{itemize}
    \item Without independently applying augmentations after cropping, the relative relationships between the cropped \textbf{parts} remain fixed.
    \item However, by performing view decoupling, the relative relationships between parts become more diverse, and we define these as \textbf{views}.
\end{itemize}

Existing self-reconstruction methods generally focus on cross-part reconstruction (e.g., block masking in Point-MAE~\cite{pang2022PointMAE}). In contrast, cross-view reconstruction (ours) significantly outperforms cross-part reconstruction, as demonstrated in the main paper Table 4, where line 4 outperforms line 1.

\section{Limitations and future work}

Point-PQAE is a novel cross-reconstruction generative learning paradigm that differs significantly from previous self-reconstruction methods, enabling more diverse and challenging pre-training. Point-MAE~\cite{pang2022PointMAE} pioneered the self-reconstruction paradigm in the point cloud self-supervised (SSL) learning field and variant optimizations are well explored, \eg, cross-modal~\cite{dong2022ACT, qi2023Recon, guo2023jointMAE}, masking strategy~\cite{zhang2023I2PMAE}, and hierarchical architecture~\cite{zhang2022pointM2AE,zhang2023I2PMAE}. \textbf{Compared to the well-studied self-reconstruction, cross-reconstruction remains significantly under-explored.} As the initial venture into cross-reconstruction, our Point-PQAE opens a new avenue for advancement in point cloud SSL. However, the model employs a vanilla transformer architecture and is constrained to single-modality knowledge. This architecture may not be optimally suited for cross-reconstruction tasks. Furthermore, the limited size of the available 3D point cloud datasets—due to the challenges in data collection—restricts the broader applicability of our single-modality approach. Future work could explore the integration of knowledge from additional modalities or the development of more efficient and appropriate architectures for the cross-reconstruction paradigm.


{
    \small
    \bibliographystyle{ieeenat_fullname}

}

\end{document}